\documentclass{article}
\usepackage{arxiv}
\usepackage{graphicx}
\usepackage[numbers]{natbib}
\usepackage{doi}
\usepackage[utf8]{inputenc} 
\usepackage[T1]{fontenc}    
\usepackage{hyperref}       
\usepackage{url}            
\usepackage{booktabs}       
\usepackage{amsfonts}       
\usepackage{nicefrac}       
\usepackage{microtype}      
\usepackage[table]{xcolor}         
\usepackage{amsmath}
\usepackage{amssymb}
\usepackage{array}
\usepackage{pifont}
\usepackage{threeparttable}
\usepackage{graphicx} 
\usepackage{array} 
\usepackage{makecell}
\usepackage{algpseudocode} 
\usepackage{multirow}
\usepackage{listings}
\usepackage{enumitem}
\usepackage{hyperref}
\usepackage{todonotes}
\usepackage{subcaption}
\usepackage{float}
\usepackage[ruled,vlined]{algorithm2e}

\newcommand{\eg}{\textit{e.g.}}

\newcommand{\ie}{\emph{i.e.}}

\usepackage{authblk}

\newcommand{\pangu}{Pangu}
\newcommand{\modelname}{\pangu~Embedded}

\title{\modelname: An Efficient Dual-system LLM Reasoner with Metacognition}

\author{\pangu~Team, Huawei\\
pangutech@huawei.com}

\renewcommand{\headeright}{Technical Report}

\begin{document}

\maketitle

\thispagestyle{fancy}

\begin{abstract}
This work presents \modelname{}, an efficient Large Language Model (LLM) reasoner developed on Ascend Neural Processing Units (NPUs), featuring flexible fast and slow thinking capabilities. \modelname{} addresses the significant computational costs and inference latency challenges prevalent in existing reasoning-optimized LLMs. We propose a two-stage training framework for its construction.
In Stage 1, the model is finetuned via an iterative distillation process, incorporating inter-iteration model merging to effectively aggregate complementary knowledge. This is followed by reinforcement learning on Ascend clusters, optimized by a latency-tolerant scheduler that combines stale synchronous parallelism with prioritized data queues. The RL process is guided by a Multi-source Adaptive Reward System (MARS), which generates dynamic, task-specific reward signals using deterministic metrics and lightweight LLM evaluators for mathematics, coding, and general problem-solving tasks.
Stage 2 introduces a dual-system framework, endowing \modelname{} with a ``fast'' mode for routine queries and a deeper ``slow'' mode for complex inference. This framework offers both manual mode switching for user control and an automatic, complexity-aware mode selection mechanism that dynamically allocates computational resources to balance latency and reasoning depth.
Experimental results on benchmarks including AIME 2024, GPQA, and LiveCodeBench demonstrate that \modelname{} with 7B parameters, outperforms similar-size models like Qwen3-8B and GLM4-9B. It delivers rapid responses and state-of-the-art reasoning quality within a single, unified model architecture, highlighting a promising direction for developing powerful yet practically deployable LLM reasoners. 
\end{abstract}

\section{Introduction}
\begin{center}
\textit{``Let your rapidity be that of the wind, your compactness that of the forest.''} \small{--- Sun Tzu}
\end{center}

Modern Large Language Models (LLMs), trained on vast textual corpora, leverage self-attention mechanisms to master intricate linguistic patterns, semantic relationships, and contextual dependencies. Over recent years, LLMs such as GPT-4~\cite{achiam2023gpt}, LLaMA~\cite{touvron2023llama}, Pangu~\cite{wang2023pangu}, and DeepSeek~\cite{liu2024deepseek} have achieved remarkable success in diverse tasks, ranging from text completion, translation, and summarization to complex applications like code synthesis~\cite{roziere2023code} and autonomous agents~\cite{wang2024survey}. Their capabilities predominantly stem from a pre-training and fine-tuning paradigm, where models first learn generalized language representations and are subsequently adapted to downstream tasks.

Beyond foundational language tasks, reasoning-optimized LLMs like OpenAI's o1~\cite{jaech2024openai} and DeepSeek-R1~\cite{guo2025deepseek} represent significant advancements, excelling in mathematical, logical, and algorithmic problem-solving. These models typically employ long chain-of-thought (CoT)~\cite{wei2022chain} processes and reinforcement learning strategies to bolster complex reasoning. For instance, DeepSeek-R1 is initialized via fine-tuning on curated CoT data and further trained with reinforcement learning that rewards rule-based output correctness, achieving impressive accuracy on math and code benchmarks. Such works demonstrate how scaling test-time computation can benefit LLM reasoning.

Despite their impressive reasoning accuracy, state-of-the-art LLM reasoners often suffer from significant deployment inefficiencies. These arise primarily from two intertwined factors: first, \textit{excessive model size} (e.g., DeepSeek-R1's 671B parameters), leading to prohibitive computational costs and memory footprints; and second, \textit{lengthy thinking processes}, where solving a single complex problem might generate over 32,000 thinking tokens, each requiring full-model inference. This linear scaling of latency with model size and reasoning steps renders them impractical for time-sensitive applications, particularly on resource-constrained devices.

To enhance the efficiency of reasoner models, recent research~\cite{feng2025efficient} has explored avenues such as shortening CoT length (e.g., via length-penalized RL~\cite{luo2025o1} or latent-space reasoning~\cite{hao2024training}), empowering smaller LLMs with strong reasoning through knowledge distillation~\cite{magister2023teaching,li2025small} or advanced RL training~\cite{zeng2025simplerl}, and accelerating decoding via efficient sampling techniques~\cite{sun2024fast,bi2024forest}.

\begin{figure} 
    \centering
    \includegraphics[width=0.8\linewidth]{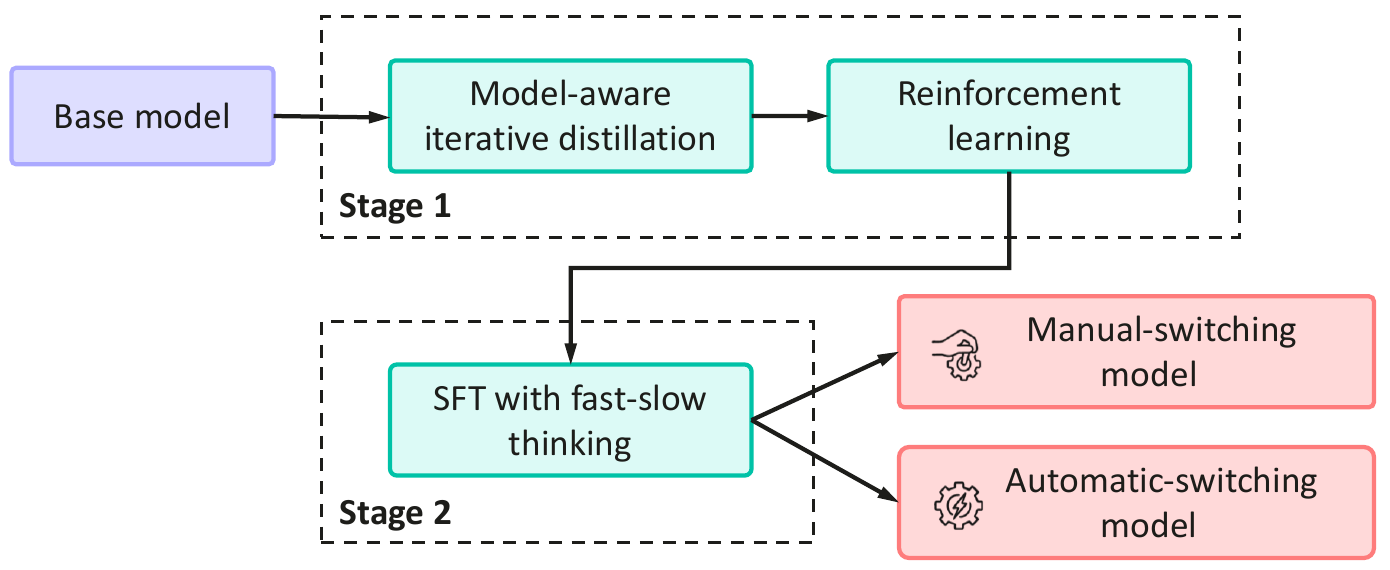} 
    \caption{An illustration of the \modelname{} training pipeline. The pipeline consists of two primary stages: Stage 1: basic reasoner construction and Stage 2: enabling fast and slow thinking in one model.}
    \label{fig:pipeline}
\end{figure}

In this work, we present \modelname{}, an efficient LLM reasoner designed to systematically address these limitations while incorporating flexible fast and slow thinking capabilities. Our core innovation is a two-stage training framework (illustrated in Figure~\ref{fig:pipeline}). Stage 1 focuses on constructing a highly capable base reasoner. This involves an iterative distillation pipeline with a two-level model merging strategy to consolidate knowledge, followed by large-scale reinforcement learning on Ascend NPU clusters. This RL phase is supported by a latency-tolerant scheduler and guided by our Multi-source Adaptive Reward System (MARS). Stage 2 then endows \modelname{} with its distinctive dual-system thinking: a ``fast'' mode for routine queries and a deeper ``slow'' mode for complex inference, offering both manual and automatic, complexity-aware mode selection. This integrated approach aims to retain broad knowledge while enhancing response quality and reliability in complex reasoning scenarios. Experimental results on various datasets show that \modelname{} with 7B parameters outperforms the similarly-sized Qwen3-8B~\cite{yang2025qwen3}, delivering both rapid responses and state-of-the-art reasoning quality within a single unified model.

\section{Basic Reasoner Construction}

The construction of our \modelname{} begins with Stage 1, dedicated to building a robust foundational reasoner. This stage, as outlined below, encompasses meticulous data preparation, a novel iterative distillation strategy, and scaled reinforcement learning tailored for Ascend NPUs.

\subsection{Preliminaries of Base Model}

\modelname{} is a 7-billion-parameter language model specifically optimized for edge computing scenarios. The pre-training data and tokenizer used for \modelname{} are consistent with those of our larger model, Pangu Ultra~\cite{yin2025pangu}. The tokenizer features a unified vocabulary of 153,376 unique tokens, ensuring balanced representation across diverse domains while maintaining overall compression efficiency. 
The pre-training corpus comprises high-quality and diverse tokens, processed in three sequential phases: the \textit{general} phase, the \textit{reasoning} phase, and the \textit{annealing} phase. 

\subsection{Post-training Data}

\begin{figure}
\centering
\includegraphics[width=1\linewidth]{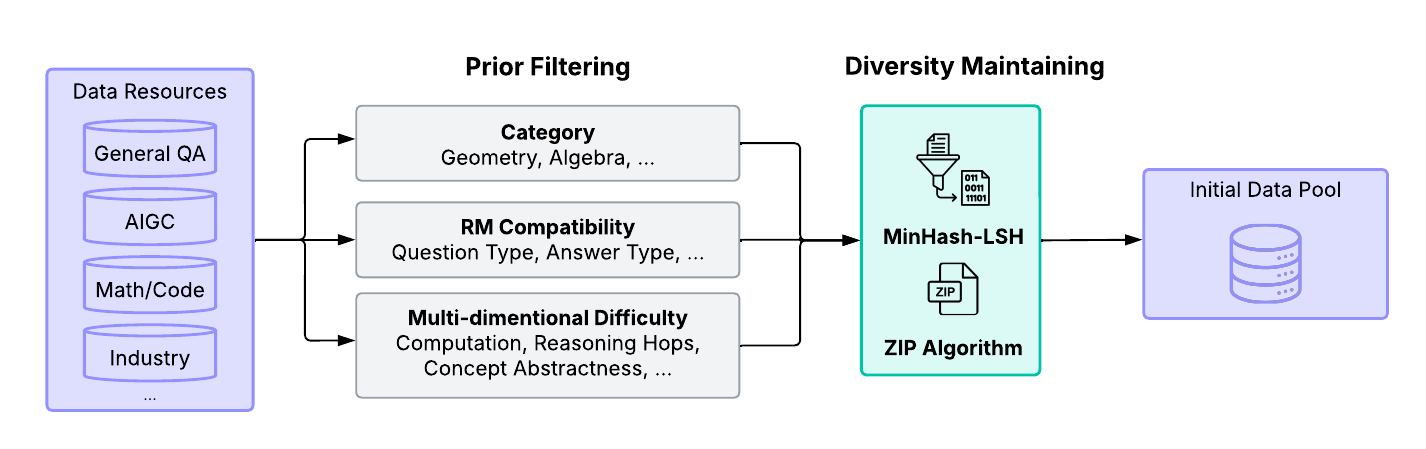}
\caption{An illustration of the construction of the initial data pool.}
\label{fig:data_pool}
\end{figure}

We first detail the construction of the initial data pool, denoted as $\mathcal{D}$, which serves as a foundational resource for subsequent development.
To assemble a high-quality and diverse set of queries, we collect data spanning reasoning and non-reasoning tasks from multiple sources. These include general Question Answering (QA), AI-Generated Content (AIGC), text analysis, coding, mathematics, logical reasoning, and tool usage, covering domains such as finance, healthcare, and government affairs. The data originate from open-sourced instructions, industrial queries, and problems synthesized from our pre-training corpora.
This collected data then undergoes a dedicated post-training processing pipeline, illustrated in Figure~\ref{fig:data_pool}, which involves two main stages: \textit{prior filtering} and \textit{diversity maintenance}.

In the \textit{prior filtering} stage, an LLM is initially used to annotate the collected data with attributes such as subcategory, \eg, geometry, algebra; question type, \eg, multiple-choice, proof question; answer verifiability, \ie, whether the answer can be verified by rules; and multi-dimensional difficulty metrics, \eg, computation complexity, reasoning hops, concept abstractness. Based on these annotations, we filter out unqualified prompts to mitigate issues like reward hacking and to reduce sampling costs. Additionally, heuristic filters are applied to remove low-quality samples and de-duplicate entries across different data sources.

To further ensure the overall \textit{diversity} of the training dataset, beyond evaluating individual samples, we focus on minimizing global redundancy. Specifically, n-gram-based MinHash-LSH is first applied to remove near-duplicate data samples. Subsequently, the final training set is selected using the ZIP algorithm, guided by entropy principles~\cite{yin2024entropy}. The ZIP algorithm directly measures instruction diversity via the data compression ratio calculated using off-the-shelf compression tools, \eg, zip. A lower compression ratio intuitively indicates higher diversity. ZIP prioritizes instruction samples with low individual compression ratios and iteratively refines the selection by adding samples that minimize similarity to already selected ones. This process yields a training subset rich in varied patterns, enabling more effective acquisition of long-CoT capabilities.

Through this rigorous process, we obtain the initial high-quality and diverse data pool $\mathcal{D}$, which is leveraged as a data source for both Reinforcement Learning (RL) and distillation phases.

\subsection{SFT Strategy: Model-aware Iterative Distillation}
\label{sec:sft}
In this section, we outline our supervised fine-tuning (SFT) strategy, with an in-depth discussion of our enhanced model-aware iterative distillation strategy.

\subsubsection{Motivation}
DeepSeek-R1~\cite{guo2025deepseek} demonstrated that superior reasoning capabilities can be effectively transferred from advanced models to smaller ones via Supervised Fine-Tuning (SFT). However, traditional SFT primarily focuses on curating high-quality, diverse instruction sets to cultivate well-rounded model capabilities. It remains less clear how to smoothly transfer advanced reasoning skills while simultaneously preserving general instruction-following abilities, especially for smaller student models.

The complexity of training data significantly influences learning outcomes, particularly for reasoning tasks. Reasoning inherently requires deeper cognitive processing compared to tasks like classification or regression. Consequently, selecting appropriate training data becomes critical, especially when distilling reasoning capabilities into more compact models~\cite{yin2025towards,li2025small}.

To investigate this, we conducted a preliminary experiment using \modelname{} on the AIME benchmark, dividing the training data into three groups based on complexity as evaluated by Equation~\eqref{eq:complexity}. The first group consisted predominantly of easy data with limited hard examples; the second featured a balanced distribution of easy and hard data; and the third comprised mostly difficult data with fewer easy samples.

\begin{table}[t] 
\centering
\caption{Average performance of \modelname{} on the AIME 2024 test set under three different data difficulty distributions.} 
\begin{tabular}{lc} 
\toprule
Data Setting & AIME Score (\%) \\ 
\midrule
No Data Selection     & 43.33 \\ 
Mostly Easy (Group 1) & 45.42 \\
Balanced (Group 2)    & 50.42 \\
Mostly Hard (Group 3) & 48.75 \\
\bottomrule
\end{tabular}
\label{tab:aime_results}
\end{table}

As shown in Table~\ref{tab:aime_results}, the balanced data group achieved the highest performance at 50.42\%, outperforming both the mostly easy (45.42\%) and mostly hard (48.75\%) groups. These skewed distributions yielded suboptimal results. The lowest performance was observed when using the full dataset without any selection, suggesting that indiscriminate data use can hinder effective learning of reasoning patterns. These findings underscore that for optimal model training in reasoning tasks, the learning difficulty of the data must be carefully tailored to the model’s current capacity. Data that is too simple may not foster necessary skill development, while overly difficult data may impede the model's ability to internalize underlying logic.

Thus, our proposed distillation pipeline prioritizes a dynamic data selection strategy, aligning the training data's complexity with the student model's evolving capabilities. This ensures the training process remains optimally challenging, fostering effective learning at each stage.

\subsubsection{Model-aware Iterative Distillation Pipeline}
\begin{figure*}[thbp] 
    \centering
    \includegraphics[width=\textwidth]{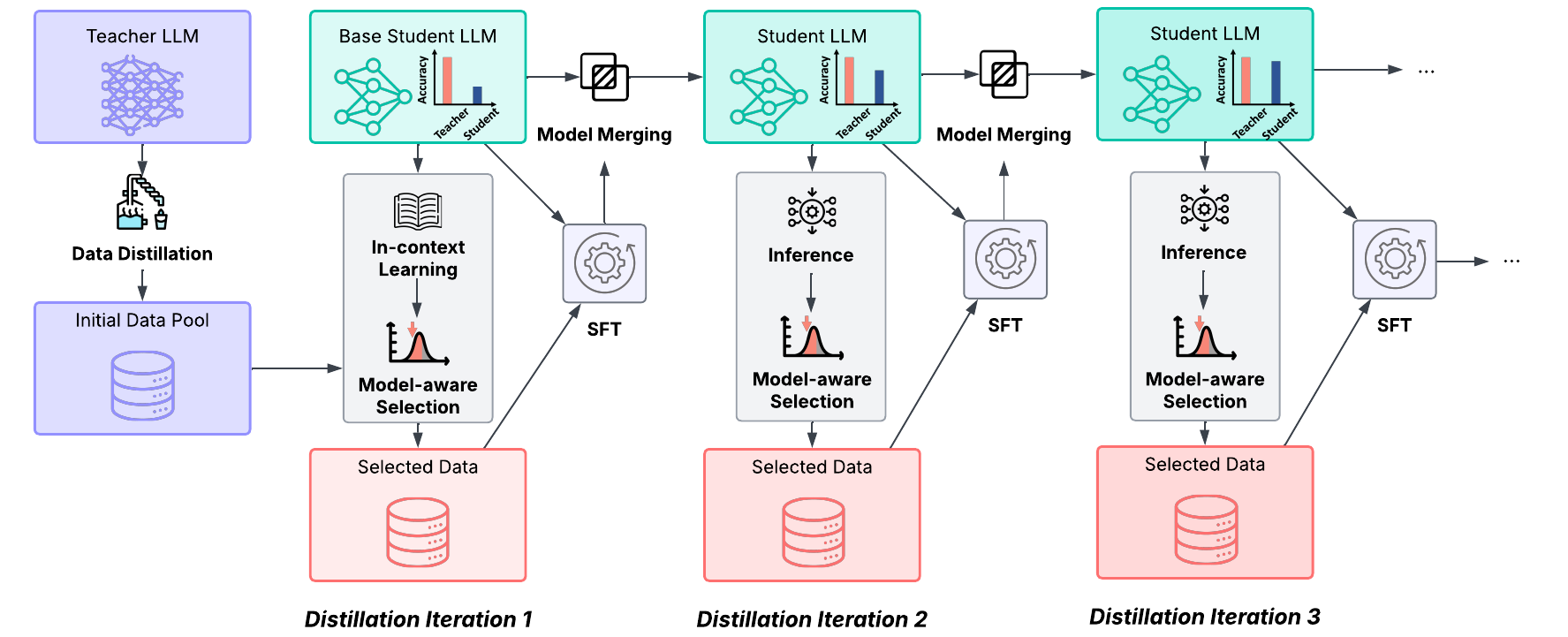} 
    \caption{The overall framework of the model-aware iterative distillation pipeline in \modelname{}. In each iteration, data samples are selectively filtered based on a model-aware complexity metric, which is evaluated using the student model from the previous iteration. This metric matches data to the student model’s current capabilities. The student model is progressively refined through multiple rounds of distillation, guided by dynamic data selection and iterative model merging.}
    \label{fig:overall}
\end{figure*}

As illustrated in Figure~\ref{fig:overall}, \modelname{} employs an iterative distillation pipeline to fully harness the knowledge from its teacher model. The process commences with the initial curated data pool $\mathcal{D}$. A model-aware data complexity metric guides this iterative process by dynamically aligning training data with the student model's evolving capabilities. The complete procedure is outlined in Algorithm~\ref{alg:iterative_training}.

\paragraph{Data Synthesis} 
The process begins by using the teacher model $\mathcal{G}_{\tau}$ to generate responses (solutions) for all queries in the initial prompt pool $\mathcal{D}_{\text{prompts}}$, thereby constructing the initial dataset for distillation $\mathcal{D}_{\text{distill}}$. This dataset is then filtered to maintain diversity and ensure quality. We implement dual-layer data verification, employing both rule-based and model-based checks, for all data samples. Specific, detailed verification is applied to mathematical and coding data, similar to the reward evaluation detailed in Section~\ref{sec:mars_rl}. 
We also emphasize rejection sampling (RS). Monte Carlo sampling can be leveraged to continuously refine generated responses towards the quality of the model's training data. Furthermore, post-selection allows us to select data based on specific preferences. We find token efficiency to be a crucial requirement for long-CoT data, which can be significantly improved by selecting the shortest correct reasoning path during RS.

\paragraph{Model-Aware Data Complexity}
Built upon the curated initial data pool $\mathcal{D}_0$, we define data complexity in a model-aware manner. For each data sample $(x, y)$ and a given student model, specifically $\mathcal{G}_{t-1}$ from the previous iteration during iteration $t$, its complexity is evaluated based on this student model's performance on that sample. The model $\mathcal{G}_{t-1}$ generates $k$ responses $\{y_1, y_2, ..., y_k\}$ for input $x$ typically using a relatively high temperature to ensure diversity. The complexity score is then defined as:
\begin{equation}
\label{eq:complexity}
C(x, y; \mathcal{G}_{t-1}) = 1 - \frac{1}{k} \sum_{i=1}^{k} \mathbb{I}(\text{Eq}(y_i,y)),
\end{equation}
where $\text{Eq}(y_1,y_2)$ examines whether the final answers in $y_1$ and $y_2$ are equal, and $\mathbb{I}(\cdot)$ is the indicator function. A lower complexity score indicates that model $\mathcal{G}_{t-1}$ already handles the data sample well, making it generally less critical for further training on that specific sample, particularly when $C(x, y; \mathcal{G}_{t-1}) = 0$. Conversely, a higher score suggests the model is still struggling. This definition is primarily applicable to samples with ground-truth answers, such as reasoning data. For non-reasoning data, complexity can be estimated using loss values.

\begin{algorithm}[htbp]
\caption{Model-aware Iterative Distillation with Inter-iteration Merging} 
\label{alg:iterative_training}
\KwIn{Pre-trained model $\mathcal{G}_{0}$ (parameters $\Theta_{\text{merged}}^{0}$), teacher model $\mathcal{G}_{\tau}$, initial training data prompts $\mathcal{D}_{\text{prompts}}$, max iterations $T$, merging weights $\lambda_1, \dots, \lambda_T$}
\KwOut{Final trained model $\mathcal{G}_{T}$ (parameters $\Theta_{\text{merged}}^{T}$)}
\tcp{Generate solutions for prompts using the teacher model}
$\mathcal{D}_{\text{distill}} \leftarrow \text{GenerateSolutions}(\mathcal{G}_{\tau}, \mathcal{D}_{\text{prompts}})$\;
\tcp{Select high-quality initial distilled data pairs (prompt, solution)}
$\mathcal{D}_0 \leftarrow \text{SelectInitialData}(\mathcal{D}_{\text{distill}})$\; 
\For{$t = 1$ \textbf{to} $T$}{
    $\mathcal{D}_s \leftarrow \{\}$\;
    \ForEach{$(x, y) \in \mathcal{D}_0$}{
        \If{$t == 1$}{
            \tcp{Add few-shot examples $\xi_{\text{fs}}$ to prompt $x$ for the first iteration}
            $x' \leftarrow \{\xi_{\text{fs}}, x\}$\; 
        } \Else {
            $x' \leftarrow x$\;
        }
        \tcp{Complexity C is evaluated using student model $\mathcal{G}_{t-1}$ (params $\Theta_{\text{merged}}^{t-1}$)}
        Compute complexity score $C(x',y; \mathcal{G}_{t-1})$ based on Equation~\eqref{eq:complexity}\;
        Compute selection probability $P(C(x',y; \mathcal{G}_{t-1}))$ based on Equation~\eqref{eq:select-prob}\;
        $p_{\text{sample}} \leftarrow \text{Uniform}(0,1)$\;
        \If{$p_{\text{sample}} < P(C(x',y; \mathcal{G}_{t-1}))$}{ 
            $\mathcal{D}_s \leftarrow \mathcal{D}_s \cup \{(x, y)\}$;
        }
    }
    \tcp{Perform SFT based on $\mathcal{G}_{t-1}$ and obtain $N_t$ checkpoints $\Theta_i^t$}
    $\{\Theta_1^t, \dots, \Theta_{N_t}^t\} \leftarrow \text{SFT\_and\_collect\_checkpoints}(\mathcal{G}_{t-1}, \mathcal{D}_s)$\;
    \tcp{Perform inter-iteration model merging using Equation~\eqref{eq:inter_iteration_merging}}
    $\Theta_{\text{merged}}^t \leftarrow \Theta_{\text{merged}}^{t-1} + \lambda_t \left( \frac{1}{N_t} \sum_{i=1}^{N_t} (\Theta_i^t - \Theta_{\text{merged}}^{t-1}) \right)$\;
    Let $\mathcal{G}_t$ be the model with parameters $\Theta_{\text{merged}}^t$\;
    \If{Performance improvement on a validation set is marginal}{
        \textbf{Break}\;
    }
}
\KwRet{$\mathcal{G}_{T}$}
\end{algorithm}

\paragraph{Iterative Training}
Our model training follows an iterative process. In each iteration $t$, the model $\mathcal{G}_{t-1}$ from the previous iteration serves as the base. The distilled training data $\mathcal{D}_s$ is adaptively selected based on this model's current capabilities, guided by the model-aware complexity score $C(x,y; \mathcal{G}_{t-1})$ defined in Equation~\eqref{eq:complexity}. The iteration begins with the pre-trained model $\mathcal{G}_0$. To enable complexity score computation for $\mathcal{G}_0$ without prior instruction tuning, we include few-shot examples $\xi_{\text{fs}}$ in the prompt for the first iteration only; these are removed in subsequent iterations.

We hypothesize that the most effective training corpus aligns with the model’s current proficiency, emphasizing data samples of medium complexity. Therefore, during each iteration, we select training samples $(x,y)$ with a probability $P(C(x,y; \mathcal{G}_{t-1}))$ related to their complexity score:
\begin{equation}
\label{eq:select-prob}
P(C) = \frac{1}{\sqrt{2 \pi \sigma^2}} \exp\left(- \frac{(C - \mu)^2}{2 \sigma^2}\right),
\end{equation}
where $C$ is the complexity score $C(x,y; \mathcal{G}_{t-1})$, and $\mu$ and $\sigma$ are predefined hyperparameters. We set $\mu$ to a value slightly less than $0.5$, as our findings suggest that incorporating more data samples of medium complexity, leaning towards simplicity, yields the best results.
As iterations progress and the model becomes more powerful, the number of low-complexity samples is expected to increase, decreasing the average complexity score. To encourage the model to tackle progressively more complex data and unlock its full potential, we keep $\mu$ constant. The iterative process stops once performance gains on a validation set become marginal.

\paragraph{Data Synthesis} 
All queries in $\mathcal{D}$ are first processed by \modelname{} to generate responses with reasoning content constructing the initial training data pool for distillation, which will be then filtered maintaining diversity.
Then, ensuring quality, we conduct dual-layer data verification, both rule-based and model-based, for all data samples, where specific detailed verification is applied to mathematical and coding data, similar to the reward evaluation in Section~\ref{sec:mars_rl}. 
In subsequent stages, data samples are iteratively selected based on complexity to better align with the evolving capabilities of the student model during fine-tuning.
We emphasize the importance of rejection sampling (RS). On one hand, we can leverage Monte Carlo sampling to continuously refine the generated response to approximate the quality of the model's training data; on the other hand, through post-selection, we can select data based on preference requirements. We find token efficiency as a crucial requirement for long-CoT data, which can be largely improved by selecting the shortest correct reasoning path during RS. 

% ...

\subsubsection{Inter-iteration Model Merging}
\label{sec:inter_iteration_model_merging} 

In the context of our iterative training approach, the data distribution for SFT dynamically evolves across training iterations, guided by the model's current proficiency. While this adaptation is beneficial, such dynamic data selection can potentially lead to issues like reduced domain coverage over time or catastrophic forgetting of previously acquired capabilities. To address these challenges and to better consolidate knowledge across iterations, we employ an inter-iteration model merging strategy. This strategy aims to smooth distributional shifts and stabilize the learning process.

Model merging has been widely used to integrate heterogeneous capabilities from models trained with different data sources or hyperparameters. Our application of model merging here, however, is tailored for an iterative refinement context. The underlying idea shares some functional similarities with the use of a reference model or KL-divergence term in reinforcement learning algorithms~\cite{schulman2017proximal,shao2024deepseekmath}, as it helps to anchor the current iteration's model to the knowledge accumulated in previous iterations.

Specifically, let $\Theta_{\text{merged}}^{t-1}$ denote the parameters of the merged model from the previous iteration $t-1$, with $\Theta_{\text{merged}}^{0}$ being the initial pre-trained model $\mathcal{G}_0$. In the current iteration $t$, after the SFT phase, we obtain $N_t$ checkpoints, denoted as $\Theta_i^t$ for $i=1, \dots, N_t$. These are typically from the last epoch of iteration $t$'s SFT. We consider $\Theta_{\text{merged}}^{t-1}$ as the reference model for this iteration's merge. We first calculate the average parametric difference, or delta, of these $N_t$ checkpoints relative to the reference model:
\begin{equation}
\bar{\delta}_t = \frac{1}{N_t} \sum_{i=1}^{N_t} (\Theta_i^t - \Theta_{\text{merged}}^{t-1}).
\label{eq:average_delta_t} 
\end{equation}
This average delta $\bar{\delta}_t$ represents the collective update learned by the SFT phase in iteration $t$. This delta is then applied to the parameters of the previous iteration's merged model, scaled by an inter-iteration merging weight $\lambda_t$. Formally, the parameters of the merged model for iteration $t$, $\Theta_{\text{merged}}^t$, are computed as:
\begin{equation}
\label{eq:inter_iteration_merging} 
\Theta_{\text{merged}}^t = \Theta_{\text{merged}}^{t-1} + \lambda_t \bar{\delta}_t = \Theta_{\text{merged}}^{t-1} + \lambda_t \left( \frac{1}{N_t} \sum_{i=1}^{N_t} (\Theta_i^t - \Theta_{\text{merged}}^{t-1}) \right).
\end{equation}
This successive delta-based merging strategy is designed to preserve knowledge from earlier stages, effectively accumulating improvements from iteration to iteration, while mitigating the risk of catastrophic forgetting. It offers a mechanism to enhance model robustness and consolidate diverse knowledge acquired in an evolving training setup. The model $\mathcal{G}_t$ in Algorithm~\ref{alg:iterative_training} corresponds to the model with parameters $\Theta_{\text{merged}}^t$.

\subsection{Enhancing Generation Quality: Repetition Self-repair}
\label{sec:repetition_self_repair} 

The supervised fine-tuning and model merging strategies yield a capable base reasoner. However, to further improve its practical usability, especially in generating long and coherent texts, we incorporate a mechanism to address potential repetition issues.
While the SFT stage produces a foundational reasoner with strong capabilities, Large Language Models can occasionally exhibit repetitive generation patterns, particularly during long text generation tasks. To mitigate this and enhance the practical generation quality of the SFT-refined model, \ie, $\mathcal{G}_t$ from Algorithm~\ref{alg:iterative_training}, we propose a Repetition Self-repair strategy. This strategy consists of two main steps: local n-gram repetition detection and prompt-controlled repetition suppression.

\paragraph{Local n-gram repetition detection method}
An n-gram is a sequence of $n$ adjacent symbols in a particular order. Traditional n-gram repetition penalties often require comparing the most recent n-gram with all preceding n-grams, leading to $O(N^2)$ computational complexity if $N$ is the sequence length. We propose a Local n-gram Detection method to improve efficiency.
First, we observe that our model tends to repeat phrases verbatim that have occurred within a local window. Detecting repetition only within this window reduces computational complexity significantly, closer to $O(N \cdot W)$ where $W$ is the window size, or even $O(N)$ if implemented efficiently. Second, we notice fewer character-level repetitions but more sentence- and paragraph-level endless loops.
Therefore, we perform detection periodically, for instance, every $t_{\text{detect}}$ decoded tokens, further reducing latency impact. We employ Jaccard Similarity to measure the similitude between two n-gram sets $A$ and $B$:
\begin{equation}
J(A, B) = \frac{|A \cap B|}{|A \cup B|}.
\label{eq:jaccard_similarity} 
\end{equation}
If this similarity for recently generated n-grams within the local window exceeds a predefined threshold, repetition is flagged.

\paragraph{Prompt-control repetition suppression strategy}
When repetition is detected through the local n-gram method, we intervene by modifying the guiding context for subsequent generation. Specifically, instead of letting the model continue its repetitive loop, we inject a control prompt designed to steer the model out of this loop. An example of such a prompt is:

\textit{Wait, I need to check if there is any repetition in the above answers. I need to check if there is any repetitive content in my answer. If there is no repetition, I will continue with the current answer. If there is any repetition, I will revise the answer to avoid redundancy.}

After inserting this control prompt, the n-gram detection window for future checks is reset or starts from the position following this inserted prompt. This mechanism encourages the model to reflect on its output and actively break the repetitive pattern.
This self-repair mechanism contributes to producing more coherent and diverse outputs from the SFT model, providing a higher quality starting point for the subsequent reinforcement learning phase.

\subsection{Scaling Reinforcement Learning on Ascend Clusters} 
\label{sec:rl_strategy_main} 

Following the supervised fine-tuning phase (Section~\ref{sec:sft}) and the incorporation of the repetition self-repair mechanism to enhance generation quality, the capabilities of the basic reasoner are further refined and aligned using large-scale RL. This section details our RL methodology, encompassing the core optimization algorithm, a sophisticated multi-source adaptive reward system (MARS), a curriculum-based approach to data mixing, and a cold-start procedure designed to effectively bootstrap the RL process.

\subsubsection{Reinforcement Learning Algorithm} 

\paragraph{Problem Formulation}
RL provides a structured framework for enhancing LLMs in the post-training stage. We formalize this via a Markov Decision Process (MDP) denoted as $\mathcal{M}=\langle \mathcal{S}, \mathcal{A}, r, T, \gamma\rangle$. The state space $\mathcal{S}$ represents the evolving context, comprising the input prompt $x_i$ and the partially generated response $y_{i,<t}$, which dynamically expands as tokens are autoregressively produced. At each step $t$, the action space $\mathcal{A}$ corresponds to selecting the next token $y_{i,t} \sim \pi_\theta(\cdot|x_i, y_{i,<t})$, inferred by the policy $\pi_\theta$ based on the current state. The transition function $T$ deterministically updates the state by appending the chosen token, reflecting the autoregressive nature of text generation. Given a labeled or reference response $y^*_i$, the reward function $r(x_i, y_i, y^*_i) \in [-1, 1]$ evaluates the complete generated response $y_i$ based on criteria such as correctness against $y^*_i$ and adherence to human preferences like coherence and safety. The discount factor $\gamma$ is set to 1. The objective of RL is to find the optimal policy $\pi^*_\theta$ that maximizes the expected reward:
\begin{equation}
\pi^*_\theta = \arg\max_{\pi_\theta \in \Pi} \mathbb{E}_{(x,y^*)\sim \mathcal{D}^*\subseteq\mathcal{D}, y\sim\pi_\theta(\cdot|x)}\left[r(x,y,y^*)\right].
\end{equation}

\paragraph{Policy Optimization with GRPO}
We employ the Group Relative Policy Optimization (GRPO) algorithm~\cite{shao2024deepseekmath}. In GRPO algirithm, for each prompt $x$, GRPO samples a group of outputs ${y_1, y_2, ..., y_G}$ from the old policy $\pi_{old}$.  The policy model $\pi_\theta$ is optimized by maximizing the following objective function $\mathcal{J}_{\text{GRPO}}(\theta)$:
\begin{equation} 
\small 
\begin{split}
&\mathcal{J}_{\text{GRPO}}(\theta) = \\&\mathbb{E}_{\substack{x \sim \mathcal{D} \\ \{y\}_{i=1}^G \sim \pi_{old}(y|x)}} \Bigg[ 
    \frac{1}{|G|}\sum_{i=1}^{|G|} \frac{1}{|y_i|}\sum_{t=1}^{|y_i|} \bigg( 
    \min\left( \frac{\pi_\theta(y_{i,t}|x,y_{i,<t})}{\pi_{\text{old}}(y_{i,t}|x,y_{i,<t})}\hat{A}_{i,t}, \text{clip}\left(\frac{\pi_\theta(y_{i,t}|x,y_{i,<t})}{\pi_{\text{old}}(y_{i,t}|x,y_{i,<t})},1-\epsilon, 1+\epsilon\right)\hat{A}_{i,t} \right) \\
    &- \beta \mathbb{D}_{\text{KL}}(\pi_\theta|| \pi_{\text{ref}}) 
\bigg) \Bigg],
\end{split}
\label{eq:grpo_objective}
\end{equation}
where $\pi_{\text{old}}$ is the policy from the previous iteration (the policy that generated the trajectories), $\pi_{\text{ref}}$ is a reference policy (often the initial SFT model) to prevent excessive deviation, $\epsilon$ is the clipping hyperparameter, and $\beta$ controls the KL divergence penalty. The estimated advantage $\hat{A}_{i,t}$ (assuming reward $r_i$ is given at the end of sequence $i$, thus $\hat{A}_{i,t}=\hat{A}_i$ for all $t$ for that sequence) is calculated using the group scores $\boldsymbol{r}=\{r_1,r_2,...,r_{G}\}$:
\begin{equation}
\hat{A}_{i} = \frac{r_{i} - \text{mean}(\boldsymbol{r})}{\text{std}(\boldsymbol{r}) + \delta_{\text{adv}}},
\label{eq:advantage_calculation}
\end{equation}
where $\delta_{\text{adv}}$ is a small constant for numerical stability.

We find that when the rewards of all responses for a single prompt are the same ($r_1=r_2=...=r_{G}$), the corresponding advantage $\hat{A}_{i}$ would be zero. In this scenario, the primary optimization signal from the PPO-clip term in Equation~\eqref{eq:grpo_objective} vanishes. Thus, the GRPO loss would decline into a behavior clone loss as $\mathcal{J}_{\text{GRPO}}(\theta)=-\beta\mathbb{D}_{\text{KL}}\left(\pi_\theta||\pi_{\text{ref}}\right)$ and it hampers the exploration of the policy.
To explicitly manage this, we proposed the \textit{Zero-Advantage-Mask}. This mechanism modifies the loss calculation such that when the advantage $\hat{A}_{i,t}$ is zero, the entire contribution of that particular sample (or token, depending on the granularity of $\hat{A}_{i,t}$) to the $\mathcal{J}_{\text{GRPO}}(\theta)$ objective becomes zero. This is formulated as follows:
Let $L_{i,t}$ be the term inside the summation for sample $i$ at step $t$ in Equation~\eqref{eq:grpo_objective}, \ie,
$L_{i,t} = \min\left( \dots \right)\hat{A}_{i,t} - \beta \mathbb{D}_{\text{KL}}(\dots)$.
Then, the effective loss term $L'_{i,t}$ used in the sum is:
\begin{equation}
L'_{i,t} = 
\begin{cases} 
L_{i,t}, & \text{if } \hat{A}_{i,t} \neq 0, \\
0, & \text{if } \hat{A}_{i,t} = 0.
\end{cases}
\label{eq:zero_advantage_mask_original_intent}
\end{equation}
This indicates that learning from samples with zero advantage is entirely skipped, preventing them from contributing to either the PPO-clip part or the KL divergence part of the loss for that specific instance. This strategy aims to focus updates on more informative samples where a clear advantage signal is present.

\subsubsection{Multi-Source Adaptive Reward System (MARS)} 
\label{sec:mars_rl} 
\begin{figure}[htb] 
    \centering
    \includegraphics[width=0.9\linewidth]{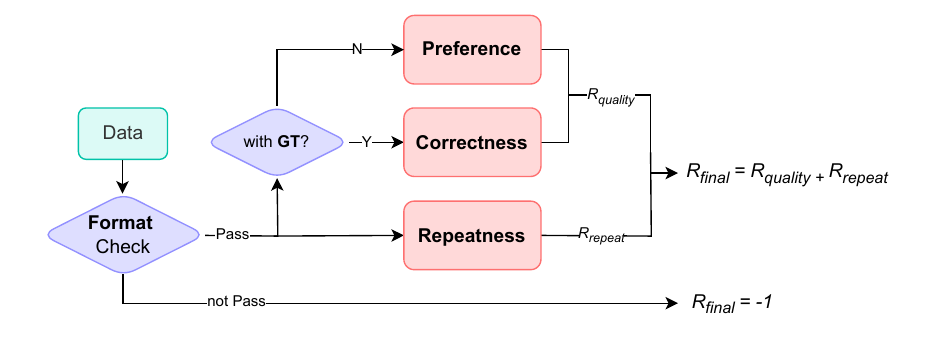} 
    \caption{An illustration of the Multi-source Adaptive Reward System (MARS).}
    \label{fig:rm_sys} 
\end{figure}

Effective alignment of reasoning-based LLMs necessitates a sophisticated reward system capable of handling task-specific verification regimes while fostering exploratory learning. Traditional single-source reward models often lack the granularity to guide complex reasoning trajectories. To address this, we propose the Multi-Source Adaptive Reward System (MARS), illustrated in Figure~\ref{fig:rm_sys}. MARS dynamically routes prompts and responses to appropriate evaluators based on predefined task labels, generating comprehensive reward signals.

\paragraph{Correctness Reward}
This reward component is primarily applied to tasks with verifiable ground truth, such as mathematics and coding problems.

Regarding mathematics problems, we employ a dual verification mechanism that integrates two key components. The first is a rule-based verifier, which handles problems where answers can be parsed and compared against ground truth using predefined arithmetic or logical rules. The second component is an LLM-based verifier; for problems requiring nuanced interpretation where rule-based methods falter, \eg, due to diverse natural language expressions of solutions, this LLM verifier assesses correctness. It takes the prompt, model response, and ground truth as input, outputting a binary judgment. Our evaluations show this LLM-based component achieves approximately 95\% precision. This hybrid approach ensures comprehensive verification across a broad spectrum of math problems.

For coding problems, rewards are computed via a detailed four-stage pipeline. This pipeline begins with code extraction from the model's response, followed by syntax verification. The third stage involves code execution using an online interpreter against provided test cases. Finally, output comparison is performed against the expected outcomes. Parallelized execution of these test cases is enabled during training to enhance efficiency. Based on the outcomes, we then utilize two distinct reward schemes. The Stage Reward scheme provides hierarchical penalties and incentives, \eg, -0.8 for syntax errors, -0.5 for complete test case mismatch, 0.1 for partial correctness, and 1.0 for full accuracy; this scheme is particularly effective for imbalanced or sparse test case scenarios. Alternatively, the Continuous Reward scheme offers a dense scoring system based on the pass rates of test cases, \eg, a linear scale from -0.5 to 0.5 for partial correctness, with 1.0 for full accuracy; this is preferred when test cases are comprehensive and quantitatively balanced.
    
\paragraph{Preference Reward} 
For open-domain data lacking verifiable ground truth, MARS incorporates a preference reward model trained to emulate human judgment. Unlike conventional RLHF pipelines that might directly use raw reward model (RM) scores, we employ a normalized preference scoring mechanism. This addresses potential inconsistencies in RM score scales across different prompts and ensures compatibility with the GRPO algorithm by stabilizing advantage estimation and preventing outlier rewards from disproportionately affecting gradient updates.

\paragraph{Other Rewards} 
To enforce desirable structural properties in generated responses beyond semantic correctness or preference, MARS includes two additional reward components, often applied as pre-processing filters or penalties:
1.  A Format Validator acts as an initial gatekeeper, penalizing or rejecting responses that violate predefined formatting requirements. (Changed from \textit{} to sentence start)
2.  A Repetition Penalty, a lightweight penalty based on Rabin-Karp string hashing, is applied to discourage undesirable repetitions, balancing creativity with conciseness. This penalty term is scaled and combined with the core reward signal. (Changed from \textit{} to sentence start)
These components operate orthogonally to the main reward channels, helping maintain output stability and structural integrity.

\subsubsection{Curriculum Data Mixing} 
\label{sec:curriculum_data_mixing} 
During Reinforcement Learning training, queries that are either excessively easy or prohibitively difficult for the current policy tend to produce constant reward signals, such as all ones for easy tasks consistently solved, or all zeros/negative values for hard tasks consistently failed. Consequently, the GRPO algorithm gains little benefit from these samples while still incurring computational overhead for rollouts and gradient computation. To address this and enhance learning efficiency, we adopt a curriculum strategy that interleaves queries of varying complexity levels, dynamically adapting to the policy's evolving capabilities.

\begin{figure}[htb] 
    \centering
    \includegraphics[width=1\linewidth]{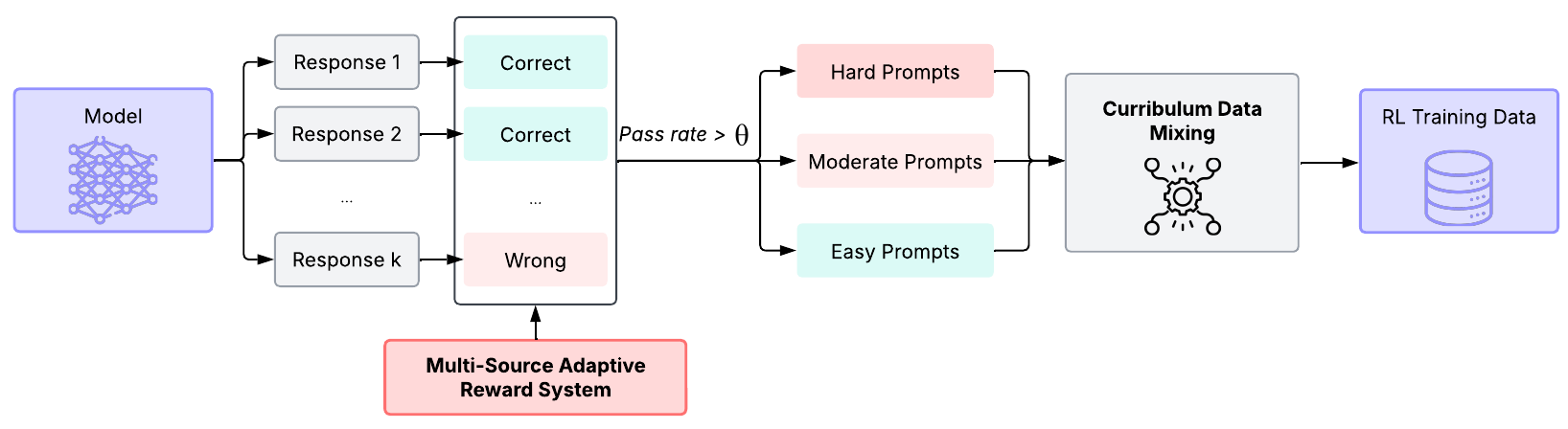} 
    \caption{An illustration of the curriculum data mixing strategy for RL training. Data complexity is assessed model-awarely, and a curated mix is fed to the RL agent.}
    \label{fig:RL_data_pipeline} 
\end{figure}

We evaluate data complexity in a model-aware manner, based on the performance of the current policy model $\pi_\theta$. Specifically, for each data sample $(x, y^*)$, where $y^*$ is the reference if available or simply $x$ if only a prompt is used for open-ended generation whose quality is assessed by MARS, the current policy $\pi_\theta$ generates $k$ responses $\{y_1, y_2, ..., y_k\}$ using a relatively high temperature to ensure response diversity. The complexity score $C(x, y^*; \pi_\theta)$ is then defined using the pass rate among these $k$ responses, conceptually similar to Equation~\eqref{eq:complexity}:
\begin{equation}
C(x, y^*; \pi_\theta) = 1 - \frac{1}{k} \sum_{i=1}^{k} \mathbb{I}(\text{Verify}(y_i, y^*)).
\label{eq:complexity_rl_context_refresh} 
\end{equation}
Here $\text{Verify}(y_i, y^*)$ is a function that checks if response $y_i$ is considered successful or high-quality with respect to the task defined by $x$ and reference $y^*$, \eg, by achieving a high score from MARS. A lower complexity score indicates that the current policy $\pi_\theta$ handles the sample well, while a higher score suggests it is still challenging. For non-reasoning or open-ended tasks without clear pass/fail criteria for individual rollouts prior to MARS evaluation, complexity might be estimated based on historical reward variance or other heuristics.

Based on this dynamic complexity assessment, as illustrated in Figure~\ref{fig:RL_data_pipeline}, each data sample is first evaluated for its complexity relative to the current policy. A curated mix of samples, balancing difficulty levels, is then progressively fed to the model. This approach helps maintain meaningful and diverse reward signals throughout training, facilitating more effective and stable policy updates.

\subsubsection{Cold Start} 
To fully unleash the potential of reinforcement learning, we implement Supervised Fine-Tuning (SFT) as a strategic warm-up phase. During this stage, a substantial seed dataset is constructed by selecting from the initial data pool $\mathcal{D}$ and further augmenting it to comprise tens of thousands of high-quality long CoT samples. We employ systematic prompt engineering and a multi-agent collaboration framework to generate CoT trajectories exhibiting considerable diversity in reasoning patterns and linguistic styles. All generated samples undergo rigorous quality assessment, including human-in-the-loop refinement, to ensure their suitability.

This meticulously prepared seed dataset empowers the SFT model, which serves as the cold-start point for RL, with several fundamental capabilities crucial for subsequent optimization. These include: (1) robust instruction following across complex task specifications; (2) an initial awareness for self-evaluation and error detection; and (3) foundational mechanisms for multi-turn reflection and iterative correction. Establishing this robust SFT baseline is critical for stabilizing the subsequent RL process and accelerating its convergence towards high-performance reasoning policies.

\subsection{RL Infrastructure on Ascend Clusters} 
\label{sec:rl_infrastructure} 

Efficient distributed training for large-scale RL is critical for enhancing LLMs with advanced reasoning capabilities. Ascend NPUs~\cite{910b2, 910b2-technical} have emerged as promising hardware alternatives for LLM training, offering strong performance and energy efficiency. However, RL pipelines are inherently more complex than SFT, involving frequent policy model rollouts, reward scoring, and parameter updates. To fully leverage Ascend hardware, a distributed RL framework optimized for Ascend NPUs is essential to maximize computational resource efficiency and accelerate end-to-end training performance.

To improve resource utilization, we support co-located training and inference of RL on the same computing cluster. The training process typically employs 8-way tensor parallelism (TP), 8-way pipeline parallelism (PP), and 16-way data parallelism (DP) on a large-scale deployment of 1,024 Ascend NPUs. Concurrently, inference tasks run on the same hardware, often configured with 8-way TP, 1-way PP, and 128-way DP, enabling effective hardware sharing without compromising performance. The global batch size is set to 256, with a rollout number of 16 per prompt.

\begin{figure}[th] 
    \centering
    \includegraphics[width=0.9\textwidth]{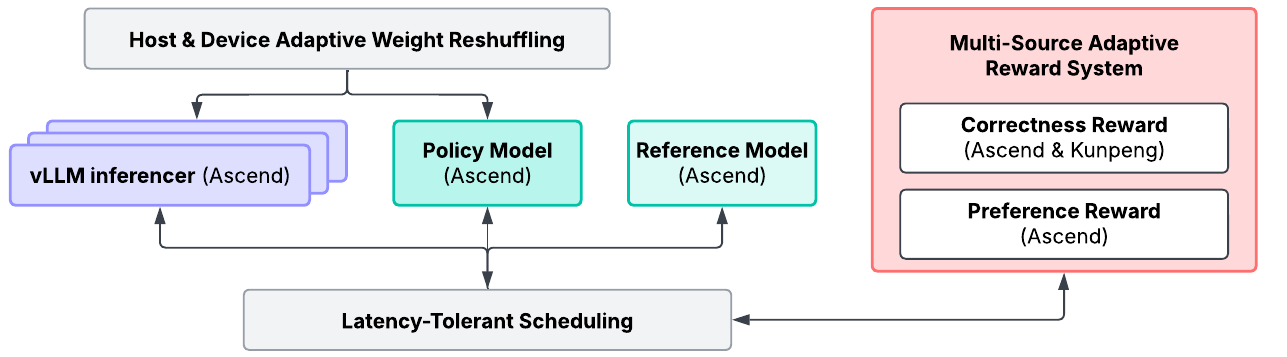} 
    \caption{Overview of the RL infrastructure on Ascend NPUs for \modelname.}
    \label{fig:system_architecture}
\end{figure}

We implement an efficient and scalable RL pipeline optimized for Ascend NPUs, as illustrated in Figure~\ref{fig:system_architecture}. The key components include: host and device adaptive weight reshuffling, the policy model, the reference model, MARS, an Ascend-optimized vLLM inferencer, and a latency-tolerant scheduling framework.

\begin{figure}[th] 
    \centering
    \includegraphics[width=\textwidth]{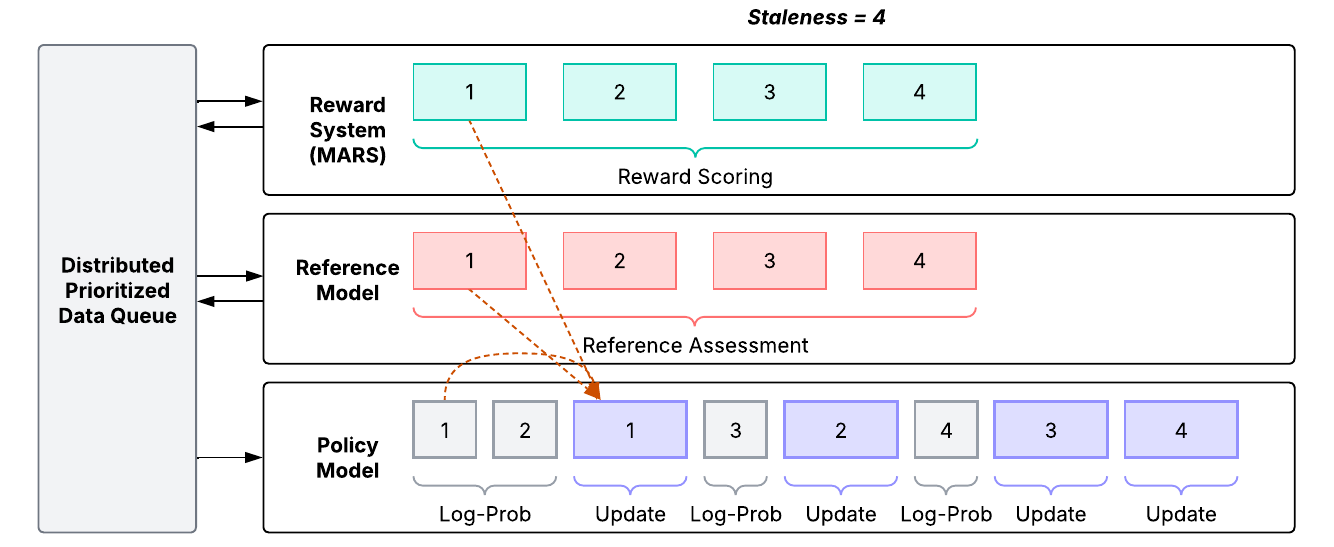} 
    \caption{The latency-tolerant scheduling framework on Ascend NPUs. Log-Prob and Update denote log-probability extraction and parameter update, respectively. Red dashed arrows indicate pipeline dependencies, where parameter update relies on the completion of reward scoring, reference assessment, and log-probability extraction for the same data.}
    \label{fig:ssp}
\end{figure}

\paragraph{Latency-tolerant scheduling framework}
We implement a latency-tolerant scheduling framework on Ascend NPUs that integrates a Stale Synchronous Parallel (SSP) scheduler~\cite{ho2013} with distributed prioritized data queues. The SSP scheduler addresses coordination bottlenecks in large-scale RL by enabling asynchronous execution with controlled staleness. As depicted in Figure~\ref{fig:ssp}, the scheduler organizes the RL pipeline into four conceptually parallel processing stages: reference assessment executed by the reference model, reward scoring by the reward model, log-probability extraction by the policy model, and gradient update also by the policy model. The parameter update stage depends on the completion of the other three.
Unlike traditional Bulk Synchronous Parallel (BSP) schemes~\cite{gerbessiotis1994direct}, where all components must synchronize before proceeding, SSP allows each pipeline stage to advance somewhat independently, within a staleness threshold typically set to $s=4$. This ensures parameter updates remain sufficiently fresh while improving overall system throughput. For instance, the log-probability extraction stage can continue processing new batches without waiting for slower components like reward scoring to complete for prior batches.
To support high-throughput execution, the scheduler incorporates a two-level distributed data queue: NPU-level queues facilitate low-latency communication within devices, while host-level queues manage cross-node coordination. On a 128-node Ascend NPU cluster, this SSP scheduling approach reduces device idle time by approximately 30\% compared to fully synchronous baselines, while maintaining training stability with less than 1\% reward drift.

\paragraph{Training-inference co-scheduling}
To maximize hardware utilization and minimize data transfer overhead, we adapt and extend concepts similar to those in HybridFlow~\cite{hybridflow2023} to support co-scheduling of training and inference on Ascend NPU clusters. Specifically, we implement host-device adaptive weight reshuffling that enables seamless sharing of model parameters between the training and inference pipelines. This mechanism supports zero-copy parameter transfers where feasible, allowing inference tasks to directly access the latest model weights without redundant memory replication or serialization. By reducing the need for dedicated hosts for each stage, co-scheduling facilitates concurrent execution of policy optimization and rollout generation on the same hardware, empirically achieving up to a 2x throughput improvement compared to isolated training and inference deployments.

\paragraph{vLLM inference optimization for Ascend NPUs}
To mitigate straggler delays caused by heterogeneous decoding lengths—which can range from 10K to over 32K tokens—we enhance the continuous batching mechanism, drawing inspiration from vLLM~\cite{kwon2023efficient,vllm2023}, with optimizations tailored for Ascend hardware. A global queue centrally manages inference requests, decoupling workload generation from NPU execution and enabling dynamic scheduling across devices. To address hardware-level memory constraints on Ascend NPUs, we introduce Ascend-aware chunking, which splits long sequences into segments aligned with memory hierarchy considerations, \eg, L2 cache sizes. Additionally, we implement proactive prefetching, overlapping token loading with attention computation by leveraging Ascend’s memory interleaving capabilities. Together, these optimizations significantly reduce latency variance across batched sequences, achieving an approximate 2x reduction in straggler-induced delays compared to static partitioning schemes, while maintaining high throughput during large-scale batched decoding.

\paragraph{Computing Clusters Configuration} 
The RL policy training and inference are deployed on a computing cluster equipped with 1,024 Ascend NPUs~\cite{910b2, 910b2-technical}. Another 256 Ascend NPUs are allocated for the reference model. The SSP scheduler and parts of the reward system are deployed on host CPUs, which utilize Kunpeng's multi-core NUMA topology. Each node in the Ascend NPU cluster houses 8 NPUs, interconnected via the Huawei Cache Coherence System (HCCS) using a full-mesh topology, and each NPU is equipped with 64GB of High-Bandwidth Memory. Inter-node communication is facilitated through RDMA over Converged Ethernet (RoCE) fabric, leveraging 200 Gbps interconnects for communication between NPUs across different nodes.

\section{Fast and Slow Thinking: A Dual System Cognitive Architecture for LLMs}
\label{sec:fast_slow_thinking_architecture}

In Stage 2, inspired by dual process theory in cognitive psychology, we introduce a dual-system framework that endows our \modelname{}, with both System 1 (fast, intuitive) and System 2 (slow, deliberative) cognitive processing capabilities. This architecture equips the model with a more nuanced and versatile reasoning toolkit. Within this framework, System 1 is characterized by direct, efficient output generation, analogous to rapid, heuristic-driven cognition. In contrast, System 2 engages in explicit, step-by-step reasoning, typically manifesting as CoT generation before deriving a final answer, mirroring methodical, analytical thought.

\begin{figure}[t] 
    \centering
    \includegraphics[width=1\linewidth]{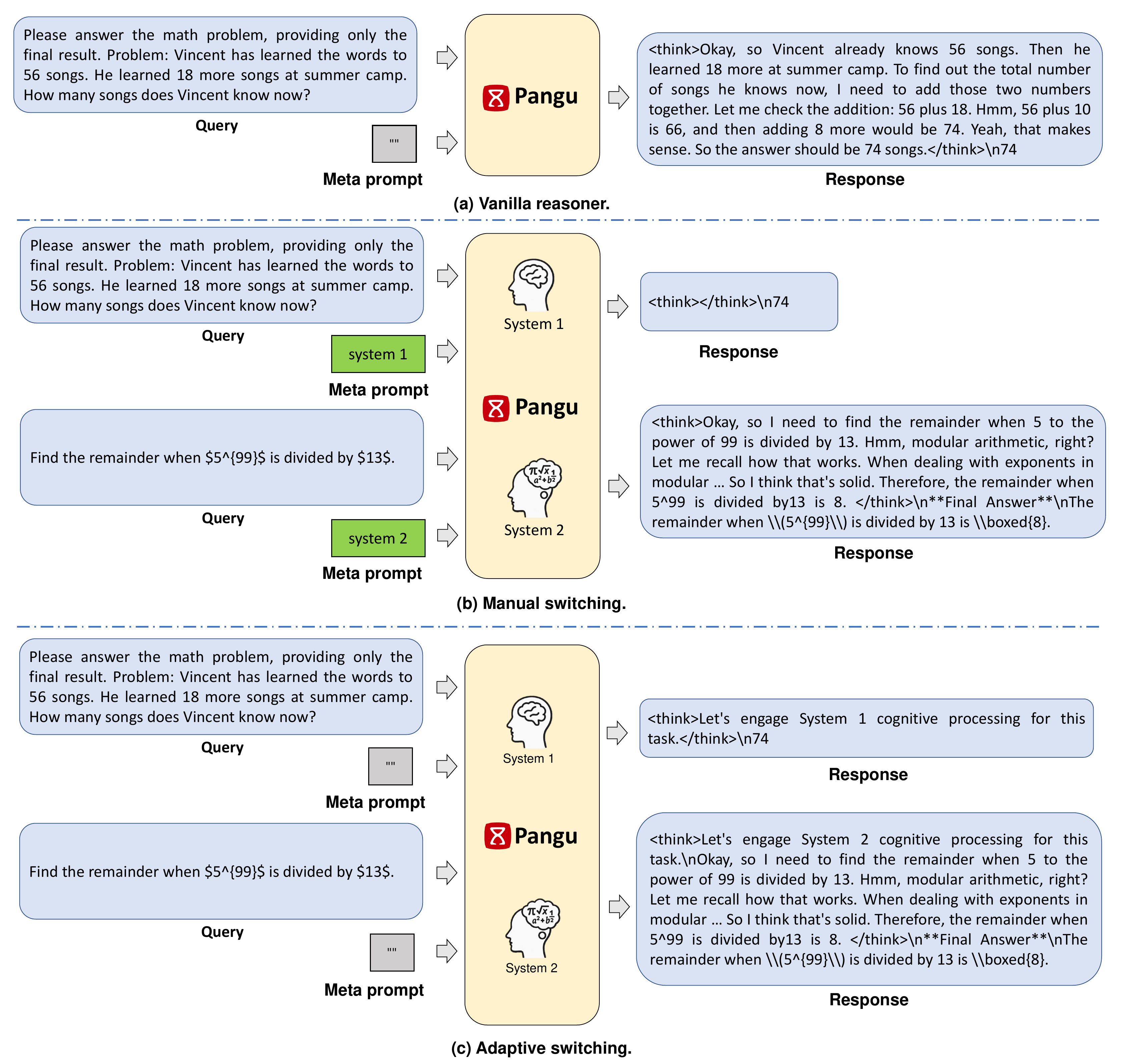} 
    \caption{Comparison of three thinking settings: 
        (a) Vanilla reasoner, typically always engaging in a default (often slow/CoT) reasoning process. 
        (b) Manual switching, where the user explicitly directs the model to use fast (System 1) or slow (System 2) thinking. 
        (c) Adaptive switching, where the model adaptively chooses the thinking mode based on assessed query complexity.}
    \label{fig:fast_slow}
\end{figure}

A key innovation of our framework, developed in Stage 2 of our overall training pipeline (see Figure~\ref{fig:pipeline}), is its support for both user-controlled manual switching between these modes and an adaptive mode selection. This section details the methodologies employed to cultivate these distinct cognitive proficiencies and interaction paradigms within a single, unified model. Figure~\ref{fig:fast_slow} provides a conceptual comparison of a vanilla reasoner versus one equipped with manual or adaptive fast-slow thinking.

\subsection{Training for Manual Mode Switching}
\label{subsec:manual_switching_training}

The manual switching capability empowers users to explicitly dictate the desired cognitive mode for \modelname{} by issuing specific directives, typically embedded within the system prompt or as special instruction prefixes. Cultivating this dual-mode proficiency within a single reasoner is achieved through a dedicated fine-tuning process. This process utilizes a curated dataset comprising exemplars of both fast thinking (System 1) and slow thinking (System 2) responses, each paired with distinct mode-specifying prompts (\eg, ``META\_PROMPT: system 1'' versus ``META\_PROMPT: system 2''). Fine-tuning the base reasoner (output of Stage 1) on this composite dataset enables it to seamlessly toggle between cognitive styles based on the user-provided directive. Specifically, we implemented a meta prompt, an independent system prompt positioned at the outset of the prompt, that explicitly instructs the model to adopt either the fast thinking mode (System 1) or the slow thinking mode (System 2). This design ensures semantic isolation from conventional system prompts, thereby avoiding interference with their intended functionality.

A critical aspect of successfully imbuing the model with these two distinct, and at times contrasting, cognitive paradigms lies in our specialized training methodology, particularly for more compact model architectures. We observed that fast thinking (direct answer generation) and slow thinking (explicit CoT reasoning) represent fundamentally different operational modes. Slow thinking, with its intricate multi-step inferential chains, is inherently more complex and challenging for models to acquire robustly. Simply co-training on both data types from scratch, or without careful sequencing, can lead to suboptimal learning or interference between the modes.

To address this, our approach leverages the proficient ``slow thinker'' model developed in Stage 1. Building upon this foundation, we undertake a crucial phase termed \textit{fusion training} for manual mode enablement. In this stage, we strategically:
\begin{enumerate}
    \item Replay the already mastered slow-thinking data. These are CoT exemplars, now prepended with the ``System 2'' directive. This replay is paramount for preserving the hard-won, nuanced slow-thinking abilities, which can be susceptible to catastrophic forgetting when exposed to a vastly different data distribution like direct fast-thinking examples.
    \item Concurrently introduce new fast-thinking data. These are direct answer exemplars (or very short CoT), prepended with the ``System 1'' directive.
\end{enumerate}
By first solidifying the more challenging slow-thinking paradigm (during Stage 1) and then carefully integrating fast-thinking capabilities through this replay-augmented fusion training, we ensure that our model retains its deep reasoning prowess while gaining the agility to respond efficiently in the fast-thinking mode when explicitly directed. This structured curriculum is instrumental in developing a truly versatile reasoner capable of adeptly navigating both cognitive pathways upon user command.
\subsection{Training for Adaptive Mode Selection}
\label{subsec:adaptive_automatic_selection_training}

In addition to user-controlled manual switching, we present a novel framework enabling \modelname{} to learn when to adaptively think fast or slow according to task complexity. As illustrated conceptually in Figure~\ref{fig:fast_slow}(c), \modelname{} can automatically output a System 1 (fast) response or a System 2 (slow) response depending on its assessment of user query complexity. This adaptive capability aims to optimize computational efficiency while maintaining high performance across tasks of varying difficulty. The methodology to achieve this involves specific data curation and a targeted fine-tuning process.

\subsubsection{Data Curation for Adaptive Learning}
\label{subsubsec:data_curation_adaptive_new}

To enable \modelname{} to discriminate task complexity and respond adaptively, we construct a specialized fine-tuning dataset, $\mathcal{D}_{\text{fusion}}$. This dataset comprises samples indicative of both fast thinking and slow thinking. For the fast thinking data subset, the response format is designed for conciseness, \eg{} a direct answer or a very brief chain-of-thought with format `{Query, CoT}'. For the slow thinking subset, the response follows a general reasoning-intensive content structure, formatted as `{Query, <think>Reasoning</think>Summary}'.

We mainly investigate the adaptive mode selection mechanism on mathematical tasks. The process to distinguish fast thinking (``easy'') and slow thinking (``hard'') queries and to construct $\mathcal{D}_{\text{fusion}}$ is as follows:
\begin{enumerate}
    \item \textbf{Complexity Evaluation}: For each question in a diverse mathematical dataset, \eg{} $\mathcal{D}_{\text{math}}$, we prompt LLMs to evaluate its computation complexity ($C_c$) and thinking complexity ($T_c$). These are scored on a scale of 1 to 5 respectively. $C_c$ refers to the computational load, such as numerical and symbolic calculations, required in the reply. $T_c$ primarily focuses on the number of reasoning steps; more steps imply higher thinking complexity.
    \item \textbf{Difficulty Classification}: Queries with $C_c \leq 2$ and $T_c \leq 2$ are labeled as ``easy'' and form the dataset $\mathcal{D}_{\text{easy}}$. The remaining queries are labeled as ``hard'', forming $\mathcal{D}_{\text{hard}}$. This heuristic approach classifies queries based on LLM-assessed intrinsic complexities.
    \item \textbf{Fast-Mode Data Generation}: For queries in $\mathcal{D}_{\text{easy}}$, we sample correct responses, typically generated by a model adept at fast, concise SFT-style answers, to construct ``fast mode'' training pairs.
    \item \textbf{Slow-Mode Data Generation}: For queries in $\mathcal{D}_{\text{hard}}$, we generate multiple detailed reasoning responses using a proficient ``slow thinking'' SFT model, such as the model resulting from our Stage 1 pipeline. Correct answers, formatted with the `<think>...</think>' structure, are sampled as ``slow mode'' training pairs.
    \item \textbf{Dataset Fusion}: We balance and merge these two subsets to form the SFT dataset $\mathcal{D}_{\text{fusion}}$, which in our experiments contained approximately 300K fast-mode and 300K slow-mode samples.
\end{enumerate}

\subsubsection{Fine-tuning for Adaptive Behavior}
\label{subsubsec:finetuning_adaptive_behavior_clarified}

To instill the adaptive fast and slow thinking capabilities, we perform a dedicated SFT stage utilizing the curated dataset $\mathcal{D}_{\text{fusion}}$ (described in Section~\ref{subsubsec:data_curation_adaptive_new}). This crucial SFT stage is initialized directly from the proficient slow-thinking reasoner developed through our comprehensive Stage 1 training pipeline. This base model, which is constructed via iterative distillation and reinforcement learning, is thus already endowed with strong general reasoning-intensive abilities.

The subsequent fine-tuning on $\mathcal{D}_{\text{fusion}}$ then specifically teaches this already capable reasoner to adapt its thinking depth according to the nature of the input query. During this process, the model learns to:
\begin{itemize}
    \item For queries associated with fast-mode responses in $\mathcal{D}_{\text{fusion}}$ (those classified as ``easy'' and formatted for concise output), generate these direct and efficient answers.
    \item For queries associated with slow-mode responses in $\mathcal{D}_{\text{fusion}}$ (those classified as ``hard'' and formatted with the `<think>...</think>' structure), replicate this structured output. This includes generating the `<think>' tag, followed by the detailed reasoning steps, and concluding with the `</think>' tag and the final summary.
\end{itemize}
By learning to generate the `<think>...</think>' block for problem types that were classified as ``hard'' (and thus formatted with these tags in $\mathcal{D}_{\text{fusion}}$), and to omit this block for ``easy'' problem types, the model implicitly learns a decision-making mechanism. The presence or absence of the `<think>...</think>' block in its own subsequent generations becomes the primary indicator of whether it has autonomously chosen a slow, deliberative reasoning path or a fast, direct one. This effectively enables the automatic switching behavior illustrated conceptually in Figure~\ref{fig:fast_slow}(c). The training objective for this SFT stage remains the standard next-token prediction loss over the sequences in $\mathcal{D}_{\text{fusion}}$. 

\subsection{User Controllability and Interaction}
\label{subsec:user_controllability}

Our \modelname{} framework, now equipped with capabilities for both explicit manual mode switching (as detailed in Section~\ref{subsec:manual_switching_training}) and implicit adaptive mode selection (trained according to Section~\ref{subsec:adaptive_automatic_selection_training}), offers a uniquely flexible interaction paradigm. While many contemporary models, \eg{} Claude 3.7, Gemini 2.5, and the Qwen3 series, incorporate hybrid fast/slow reasoning through user-controlled switches that typically require explicit API parameters or specific prompt-based suffixes, our approach emphasizes a more integrated and natural user experience.

By default, \modelname{} leverages its trained adaptive reasoning, autonomously selecting the appropriate reasoning depth based on its assessment of problem difficulty and category, especially when no explicit user preference for a mode is given. However, a crucial aspect of our system is its ``user-in-the-loop'' flexibility. Even when the model operates adaptively, users can override its autonomous decisions or guide its response style by issuing explicit natural language instructions. For instance, a user might provide feedback such as, ``Your answer is too complicated, please try a simpler one''.

This design offers soft, intuitive control over the model's reasoning style. It also anticipates the future direction of human-AI interaction, where voice and natural language are increasingly likely to become dominant interface modalities. Compared to existing parameter-driven or rigid prompt-based mode switching, our method prioritizes both adaptability and a seamless user experience, positioning it well for practical deployment in interactive scenarios.

\begin{figure}[htb] 
    \centering
    \includegraphics[width=0.98\linewidth]{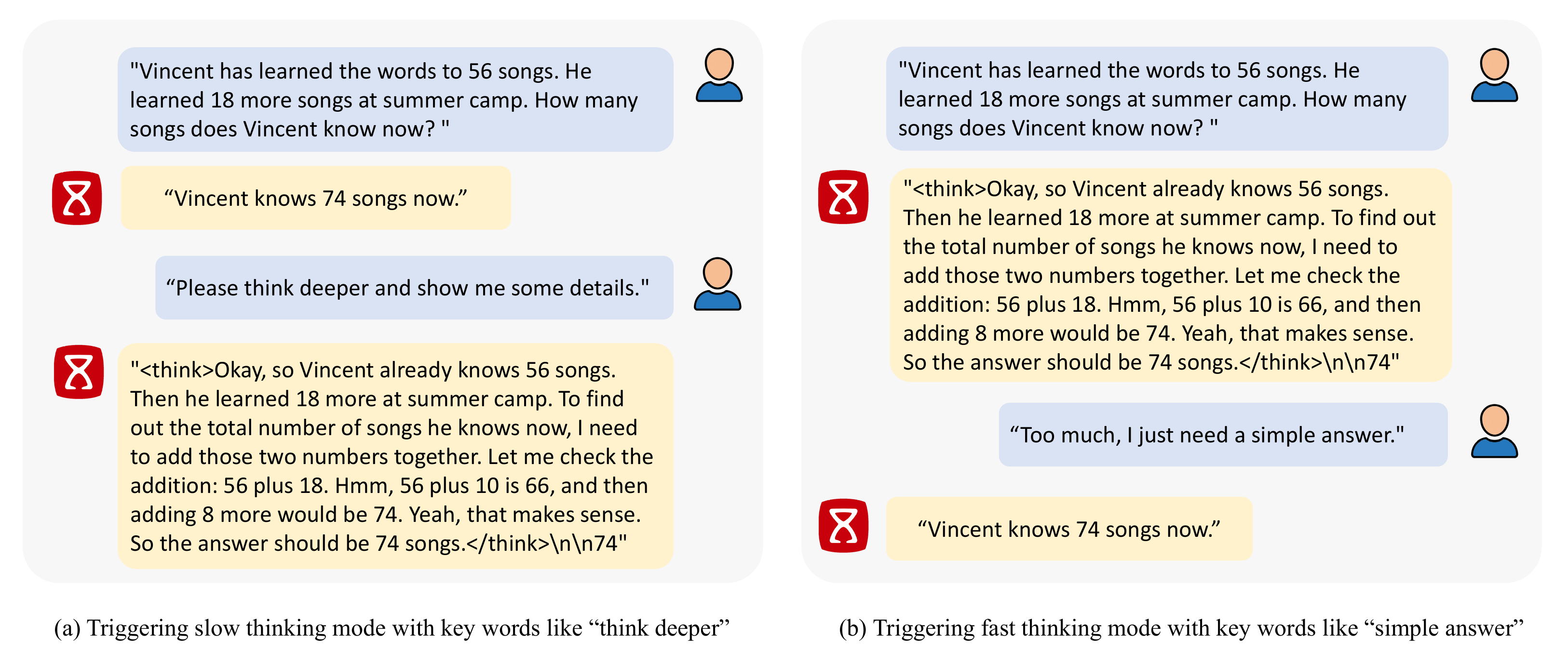} 
    \caption{An illustrative example of fast and slow adaptive thinking switch under user command with natural language context. The model initially provides a detailed (slow thinking) response, and then, upon user request for a simpler answer, switches to a concise (fast thinking) response.}
    \label{fig:multi-turn-example}
\end{figure}

Figure~\ref{fig:multi-turn-example} illustrates a typical user-model interaction scenario enabled by this natural language controllability. In this example, the user first poses a math problem and receives a detailed, step-by-step solution generated by the model's slow thinking mode (System 2). The user then requests a simpler, more concise answer using a natural language instruction. In response, the model dynamically switches to its fast thinking mode (System 1) and provides a brief, direct solution. This example highlights the flexibility and intuitiveness of our system: users can effortlessly influence the model's reasoning style in real time using natural language, without needing to manage explicit parameters or engage in complex prompt engineering. Such interaction not only improves user experience but also paves the way for more natural, potentially voice-driven, human-AI collaboration in future applications.

\section{Experiments}
\label{sec:experiments} 

In this section, we evaluate the performance of \modelname{} on a diverse set of benchmarks, encompassing both complex reasoning and general language understanding tasks. We detail our training configurations, main experimental results, and extensive ablation studies that validate our design choices.

\subsection{Training Setup}
\label{subsec:training_setup}

\subsubsection{Iterative Distillation (SFT) Training Setup}
\label{ssubsec:distillation_training_setup}

For the SFT stages within our model-aware iterative distillation pipeline (detailed in Section~\ref{sec:inter_iteration_model_merging} and Algorithm~\ref{alg:iterative_training}), \modelname{} models are typically trained for 6 epochs in each iteration, using a global batch size of 64. 
The learning rate follows a cosine decay schedule, starting at $5 \times 10^{-6}$ in the first SFT iteration and gradually decreasing to 10\% of its peak value over the training period of that iteration. We utilize the AdamW optimizer with a weight decay of 0.1 and apply gradient clipping at a norm of 1.0 to stabilize training. 
In each subsequent SFT iteration $t$, the model is initialized from the merged checkpoint $\mathcal{G}_{t-1}$ (with parameters $\Theta_{\text{merged}}^{t-1}$) from the previous iteration. A brief warm-up phase is performed, with the learning rate determined by the cosine decay schedule, ensuring it does not exceed the peak learning rate of the previous iteration. Our experiments indicate that an excessively high learning rate, especially one exceeding the starting learning rate of the prior iteration, significantly impairs model performance. This is primarily because as the model tackles increasingly difficult tasks or refines its capabilities on more nuanced data distributions selected by the model-aware complexity metric, a smaller learning rate enables more stable convergence and allows the model to better capture subtle solution patterns, thereby unlocking greater potential.
To optimize hardware utilization during these SFT stages, we set the maximum sequence length to 32,768 tokens. Multiple samples are concatenated to form sequences that fully utilize the available computational resources.

\subsubsection{RL Training Setup}
\label{ssubsec:rl_training_setup}

\paragraph{Hyperparameter Search}
To determine optimal training parameters for the RL phase, we conducted a series of hyperparameter search experiments. Specifically, we performed a grid search focusing on key hyperparameters including mini-batch size, sampling temperature, and the KL loss coefficient.

The mini-batch size critically influences the number of updates per training step, thereby affecting the balance between training stability and sample efficiency. Our empirical results indicated that when the model is updated more than twice per step with the same batch of experience, a rapid decrease in reward is often observed after a few training steps, and the model's responses tend to become more random or less coherent.

The sampling temperature directly impacts the diversity of generated responses and, consequently, the format-related reward components. We tested various sampling temperatures. High temperatures, \eg, 1.0, initially led to low format rewards; although the model could learn to adhere to format constraints within a few steps, its outputs often suffered from poor readability and undesirable language mixing as training progressed. Conversely, low sampling temperatures, \eg, below 0.6, significantly reduced response diversity. The pass rate of multiple responses generated for the same prompt declined sharply, leading to insufficient exploration during the RL training phase.

The KL loss coefficient also plays a crucial role in model training and performance optimization. An excessively large KL coefficient tended to make the model oscillate around its initial policy (the starting point of RL), hindering meaningful learning progress. Conversely, removing the KL coefficient entirely sometimes led to rapid initial improvement, but often resulted in a significant drop in accuracy after several training steps due to excessive deviation from the reference policy.

\paragraph{Training Details}
Based on our hyperparameter search and empirical tuning, the following settings were adopted for the RL training phase. In each training step, 512 prompts are sampled from the curated dataset described in Section~\ref{sec:rl_strategy_main} (referring to data mixing). For each prompt, 8 responses are sampled to estimate advantages. We set the approximate ratio of easy, medium, and hard prompts to 1:7:2. This distribution aims to ensure that the normalized advantages for most samples are non-zero, providing effective learning signals. The learning rate is set to $1 \times 10^{-6}$. The optimal configuration for other key parameters included a mini-batch size of $256 \times 8$ prompt-response pairs processed effectively by the policy, a default sampling temperature of 0.9, a KL loss coefficient of $1 \times 10^{-2}$. For the clipping ratio $\epsilon$, we adopt the clip higher configuration of 0.28 from DAPO~\cite{yu2025dapo}.

\subsection{Main Results}
\label{subsec:main_results}

\paragraph{Evaluation Benchmarks.}
We conduct a comprehensive evaluation of \modelname{}'s capabilities across several domains. These include:
\begin{itemize}[leftmargin=*,nosep]
    \item Sophisticated reasoning tasks, encompassing mathematical competence measured by AIME 2024~\cite{MAA} and MATH-500~\cite{lightman2023let}; coding competition benchmarks, specifically LiveCodeBench~\cite{jain2024livecodebench}; and scientific reasoning tasks, such as GPQA Diamond~\cite{rein2024gpqa}.
    \item General language comprehension and reasoning capabilities, represented by MMLU-Pro~\cite{gema2024we} and Arena Hard~\cite{arenahard2024}.
\end{itemize}

\paragraph{Evaluation Baselines and Metrics.}
In our main comparative analysis (Table~\ref{tab:MyModelComparisonGroupedFinalFormatted}), we evaluate the performance of our 7B parameter model, \modelname{} (which integrates fast and slow thinking capabilities into a single model), against contemporary models of similar scale. These include Qwen3-8B, GLM-4-9B, and Nemotron-Nano-8B. We primarily use Pass@1 accuracy as the evaluation metric.

\paragraph{Decoding Strategy.}
We observed that using greedy decoding to evaluate reasoning models, especially those employing long thinking processes, often results in repetitive content. To mitigate this issue in \modelname{}, our decoding strategy first applies top-$n\sigma$ sampling~\cite{tang2024topnsigmalogitsneed}, which operates directly on pre-softmax logits by leveraging a statistical threshold. Following this, we employ top-$p$ sampling, succeeded by random token selection if further diversification is needed. This multi-stage decoding strategy effectively reduces repetition rates and enhances overall generation quality and performance. To ensure stable and reliable results, particularly on smaller benchmarks, we set a minimum evaluation threshold of 500 effective samples per benchmark. Let $M$ denote the benchmark size; the number of evaluation runs $N$ is determined as the minimal integer satisfying:
\begin{equation}
    N = \min \left\{ n \in \mathbb{N} \mid n \times M \geq 500 \right\}.
\end{equation}

\paragraph{Main Comparison Settings.}
The primary comparison of \modelname{} (7B, unified model) against Qwen3-8B, GLM-4-9B, and Nemotron-Nano-8B is detailed in Table~\ref{tab:MyModelComparisonGroupedFinalFormatted}. For these evaluations, the majority of our assessments employ few-shot inputs, with a minority using zero-shot prompts, aligning with standard practices for each benchmark. We evaluate most benchmarks with gold answers through exact matching, while employing execution-based verification for code generation tasks.

\begin{table}[tb] 
    \centering
    \renewcommand{\arraystretch}{1.2} 
    \caption{Comparison of different models across benchmarks. \modelname{} results are for the 7B unified model. ``Nothinking (system1)'' and ``Thinking (system2)'' correspond to its fast and slow thinking modes, respectively. $\dagger$ indicates results from the original report, otherwise are evaluated by our internal evaluation. The best results in each mode are in \textbf{Bold}.} 
    \label{tab:MyModelComparisonGroupedFinalFormatted}
    \resizebox{\linewidth}{!}{ 
    \begin{tabular}{@{}llccccc@{}}
    \toprule
    \multirow{2}{*}{\textbf{Model}} & \multirow{2}{*}{\textbf{Mode}}                       & \multirow{2}{*}{\textbf{GPQA}} & \multirow{2}{*}{\textbf{AIME24}} & {\textbf{LiveCodeBench}} & \multirow{2}{*}{\textbf{ArenaHard}} & \multirow{2}{*}{\textbf{MMLU-Pro}} \\ 
    &&&&(24.08-25.01)&&\\
    \midrule
    Qwen3-8B            & \multirow{4}{*}{Nothinking (system1)} & 39.3$^\dagger$         & {29.1}   & 24.5                   & \textbf{79.6}$^\dagger$      & 62.2             \\
    GLM-4-9B            &                                     & 47.0          & 6.7             & 18.4                   & 54.7               & 60.1              \\
    Nemotron-Nano-8B   &                             & 39.4$^\dagger$          & 3.0$^\dagger$             & 19.5          & 16.2               & 34.6         \\
    \modelname          &                                     & \textbf{58.0} & \textbf{35.8}   & \textbf{27.2}                   & 77.0               & \textbf{67.2}     \\ 
    \midrule
    Qwen3-8B            & \multirow{4}{*}{Thinking (system2)}   & 62.0$^\dagger$         & 79.4            & 61.8                   & {89.9}      & 72.5              \\
    GLM-4-9B            &                                     & 58.5$^\dagger$          & 76.4$^\dagger$           & 51.8$^\dagger$                  & 67.4$^\dagger$              & 72.4              \\
    Nemotron-Nano-8B  &                             & 54.1$^\dagger$          & 61.3$^\dagger$            & 51.4                  & 23.5            & 42.4             \\
    \modelname          &                                     & \textbf{68.0} & \textbf{81.9}   & \textbf{67.1}          & \textbf{93.9}      & \textbf{79.0}     \\ 
    \bottomrule
    \end{tabular}
    }
\end{table}

As shown in Table~\ref{tab:MyModelComparisonGroupedFinalFormatted}, our unified \modelname{} model exhibits strong and often superior performance across multiple benchmarks when operating in its distinct thinking modes.
When employing its ``Thinking (system2)'' mode, \modelname{} demonstrates leading capabilities on several reasoning-intensive benchmarks. In its ``Nothinking (system1)'' mode, designed for efficiency, \modelname{} remains highly competitive. These results underscore the effectiveness of our single \modelname{} model. Its integrated fast and slow thinking architecture allows it to achieve both top-tier reasoning performance in its deliberative ``Thinking'' mode and strong, efficient performance in its direct ``Nothinking'' mode, often leading the comparison group of similarly sized models.

\subsection{Ablation Studies on Manual Switching}
\label{subsec:results_manual_switching_ablation}

To rigorously evaluate our approach for enabling manual fast and slow thinking modes within a single \modelname{} model, we conducted ablation studies focusing on two primary aspects. First, we compare the performance of our unified \modelname{}, when manually switched to a specific mode, against specialist models trained exclusively for either fast or slow thinking. Second, we demonstrate the superiority of our specialized fusion training strategy (detailed in Section~\ref{subsec:manual_switching_training}) over a naive co-training baseline for instilling these dual-mode capabilities. Table~\ref{tab:ablation_study_comprehensive} presents these comparisons. Our textual analysis highlights overall trends observed across the benchmarks.

\begin{table}[h]
    \centering
    \caption{Ablation study results for manual switching across benchmarks. ``Sys1'' denotes the fast thinking mode (System 1), and ``Sys2'' denotes the slow thinking mode (System 2). \modelname{} refers to our unified model with manual switching. Specialist models were trained only for a single mode.}
    \label{tab:ablation_study_comprehensive}
    \resizebox{\linewidth}{!}{ 
    \begin{tabular}{@{}llccccc@{}}
    \toprule
    \textbf{Model Configuration} & \textbf{Mode} & \textbf{GPQA} & \textbf{AIME24} & \textbf{LiveCodeBench} & \textbf{ArenaHard} & \textbf{MMLU-Pro} \\
    \midrule
    \multirow{2}{*}{\modelname{} } & Slow (Sys2) & 68.0 & 81.9 & 67.1 & 93.9 & 79.0 \\ 
                                       & Fast (Sys1) & 58.0 & 35.8 & 27.2 & 77.0 & 67.2 \\ 
    \midrule
    Specialist: Slow Only                & Slow (Sys2) & 72.2 & 81.8& 66.7 & 89.7 & 80.9 \\ 
    Specialist: Fast Only                & Fast (Sys1) & 55.6 & 28.6 & 26.1 & 79.9 & 70.9 \\ 
    \midrule
    \multirow{2}{*}{Naive Co-training}   & Slow (Sys2) & 65.1 & 70.2 & 50.5 & 82.8 & 70.4 \\ 
                                       & Fast (Sys1) & 50.4 & 26.9 & 24.7 & 75.2 & 61.8 \\ 
    \bottomrule
    \end{tabular}
    }
\end{table}

As detailed in Table~\ref{tab:ablation_study_comprehensive}, our unified \modelname{} model, when manually directed, demonstrates strong performance in both its thinking modes, presenting an interesting profile when compared against specialist models.
In its slow thinking mode (System 2), the unified \modelname{} shows highly competitive results. While the ``Specialist: Slow Only'' model exhibits a lead on certain benchmarks like GPQA and MMLU-Pro, our unified model achieves comparable or even slightly better performance on other demanding tasks such as AIME24 and LiveCodeBench, and shows a notable advantage on ArenaHard. This suggests that the unified model largely preserves strong deliberative reasoning capabilities.

When switched to its fast thinking mode (System 1), the unified \modelname{} again proves effective. Interestingly, it surpasses the ``Specialist: Fast Only'' model on some key reasoning benchmarks, particularly GPQA and AIME24. On other benchmarks like LiveCodeBench, ArenaHard, and MMLU-Pro, the specialist fast model maintains an edge. The strong performance of the unified model in fast mode on complex reasoning tasks indicates that our fusion training approach successfully cultivates robust rapid reasoning, potentially benefiting from the comprehensive knowledge integrated during the overall training process.
The ability of a single, unified \modelname{} to deliver such compelling dual-mode performance, closely rivaling and at times outperforming specialist counterparts, underscores the effectiveness of our approach in creating a versatile yet efficient model.

A cornerstone of enabling these dual manual modes is our specialized fusion training strategy, detailed in Section~\ref{subsec:manual_switching_training}. This strategy incorporates replay of mastered slow thinking data while introducing fast thinking exemplars. To validate its critical importance, we compare the unified \modelname{} with a ``Naive Co-training'' baseline, which was trained by simply mixing all fast and slow thinking data without the structured curriculum and replay mechanisms of our fusion training.
The results in Table~\ref{tab:ablation_study_comprehensive} unequivocally demonstrate the substantial advantages of our fusion training methodology. The ``Naive Co-training'' model consistently underperforms our unified \modelname{} across all benchmarks in both its slow and fast thinking modes. The performance gap is often considerable, particularly on more complex reasoning tasks. This highlights that a sophisticated training strategy, such as our fusion approach, is indispensable for effectively teaching a model distinct operational modes while mitigating mutual interference and preventing significant performance degradation.

\begin{figure}[h!]
\centering
\includegraphics[width=1\linewidth]{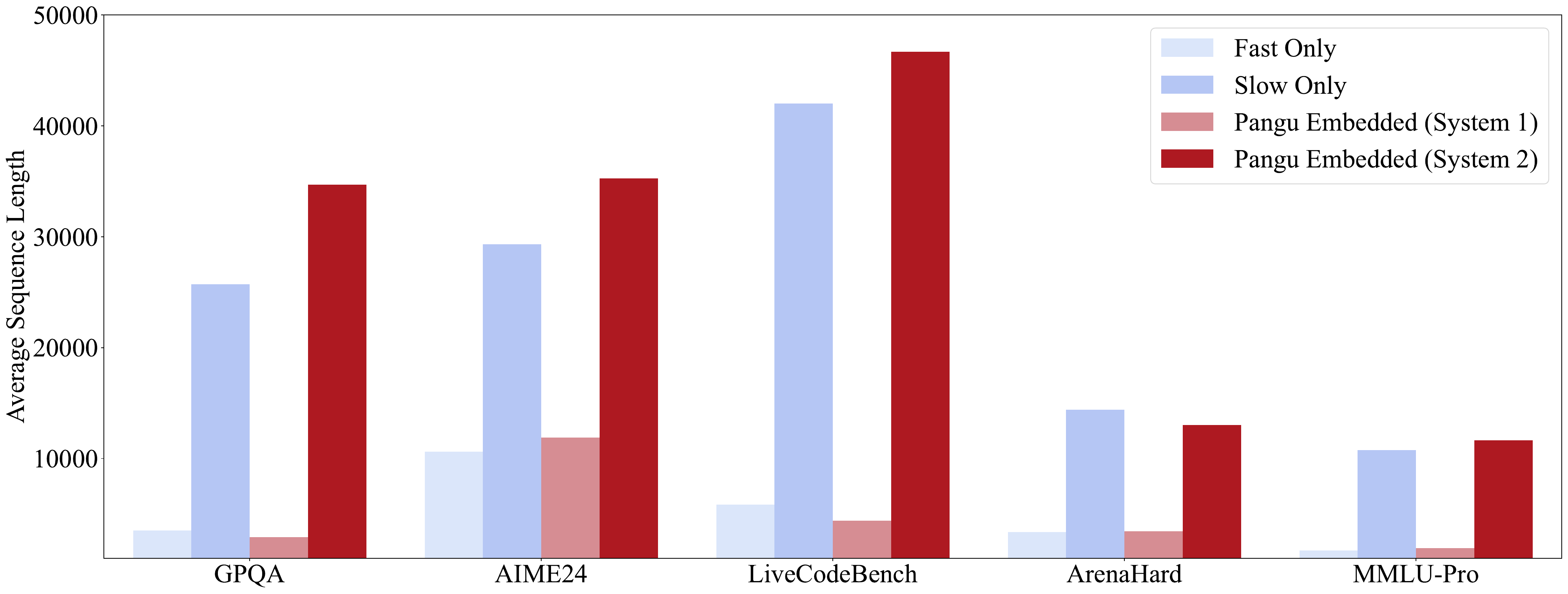}  \caption{Average sequence length of responses generated by different model configurations across various benchmarks. ``Fast Only'' and ``Slow Only'' represent specialized models. ``Pangu Embedded (System 1)'' corresponds to the fast thinking mode of our model, and ``Pangu Embedded (System 2)'' corresponds to the slow thinking mode.}
\label{fig:sequence_length_comparison}
\end{figure}

In addition to performance metrics, we also examined the average sequence length of the generated responses, as depicted in Figure~\ref{fig:sequence_length_comparison}. A clear observation is that the slow thinking modes, both in our \modelname{} (System 2) and the specialized ``Slow Only'' model, consistently produce significantly longer sequences compared to their fast thinking counterparts (``Pangu Embedded (System 1)'' and ``Fast Only'') across all benchmarks. This aligns with the expectation that slow thinking, which often involves multi-step reasoning and detailed elaboration, naturally requires more tokens to articulate. For instance, on GPQA, AIME24, and LiveCodeBench, the average sequence lengths for System 2 are substantially higher than for System 1. This distinction is less pronounced but still present for ArenaHard and MMLU-Pro, which may contain a higher proportion of questions answerable with more concise responses even in a slow thinking paradigm.

Interestingly, the average sequence lengths for our \modelname{} in System 1 and System 2 modes generally mirror those of the ``Fast Only'' and ``Slow Only'' specialist models, respectively. For example, the ``Pangu Embedded (System 1)'' lengths are very close to the ``Fast Only'' lengths across all benchmarks. Similarly, ``Pangu Embedded (System 2)'' sequence lengths are comparable to, and in some cases like GPQA, AIME24 and LiveCodeBench, even exceed those of the ``Slow Only'' model. This suggests that our fusion-trained model effectively adopts the characteristic verbosity (or conciseness) associated with each thinking style, rather than averaging out or exhibiting an unnatural length profile. There isn't a single, striking conclusion from these length observations other than the expected pattern of longer outputs for System 2 and the fact that our dual-mode model behaves comparably to specialized models in this regard.

In summary, these ablation studies validate two key aspects of our work. First, the manual switching mechanism within \modelname~successfully allows a single model to achieve performance levels akin to those of separate, specialized fast thinking and slow thinking models across a diverse set of benchmarks. Second, our specific fusion training strategy, characterized by its careful sequencing and replay of training data, is demonstrably superior to naive co training approaches and is essential for effectively instilling robust dual mode cognitive capabilities in the model.

In summary, these ablation studies validate two key aspects of our work regarding manual controllability. First, the manual switching mechanism within the unified \modelname{} allows a single model to achieve highly competitive performance levels in both fast and slow thinking modes, broadly comparable and in some instances superior to dedicated specialist models. Second, our specific fusion training strategy is demonstrably and significantly superior to naive co-training approaches, proving essential for effectively instilling these versatile, user-controllable dual-mode capabilities within one efficient model architecture.

\subsection{Ablation Studies on Adaptive Fast and Slow Thinking}
\label{sec:exp-setup}

In this subsection, we report the accuracy and average token usage of \modelname{} when operating in its adaptive fast and slow thinking mode. The average token count reported for all benchmarks is calculated as the mean within the 95th percentile of token lengths for generated responses, a measure intended to reduce the impact of extreme outliers on the average.

\begin{table}[tb] 
\centering
\renewcommand{\arraystretch}{1.2}
\caption{Accuracy (\%) and Average Token Usage for \modelname{} (Baseline) versus \modelname{} (Adaptive) across key datasets. The average token count is computed as the mean within the 95th percentile.}
\label{tab:acc-token}
\begin{tabular}{@{}lcccc@{}} 
\toprule
\multirow{2}{*}{\textbf{Model}} & \multicolumn{2}{c}{\textbf{MATH500}} & \multicolumn{2}{c}{\textbf{GSM8K}} \\
\cmidrule(lr){2-3} \cmidrule(lr){4-5} 
 & Acc. & Avg. Token & Acc. & Avg. Token \\ 
\midrule
\modelname{} (Baseline)  & 92.4  & 3,066 & 95.98 & 2,721 \\
\modelname{} (Adaptive)  & 92.2  & 2,725 & 95.39 & 325   \\
\bottomrule
\end{tabular}
\end{table}

We compare the accuracy and average token usage of the \modelname{} (Baseline) model against the \modelname{} (Adaptive) model, which incorporates the adaptive fast and slow thinking mechanism, on the MATH500 and GSM8K benchmarks. As shown in Table~\ref{tab:acc-token}, while the accuracy remains nearly unchanged across these datasets—a crucial indicator of preserved reasoning quality—the \modelname{} (Adaptive) model demonstrates a significant reduction in average token usage. This reduction is approximately 11\% on MATH500 and a substantial 88\% on GSM8K. The results suggest that the reduction in average token count is more pronounced for datasets generally considered simpler, such as GSM8K, where faster, more direct responses are often sufficient.

\begin{figure}[!t] 
    \centering
    \includegraphics[width=0.6\linewidth]{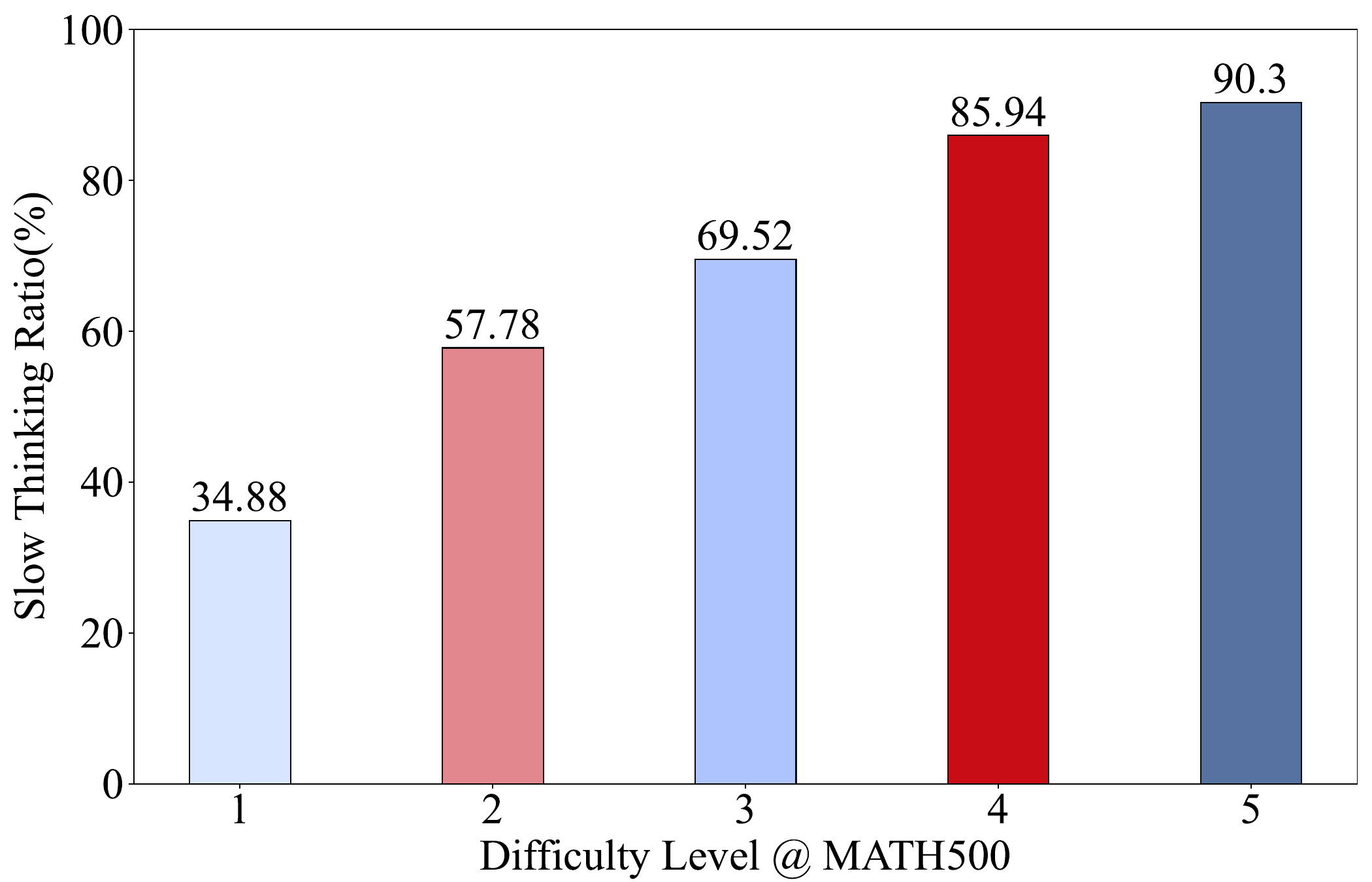} 
    \caption{Analysis of the proportion of queries for which the slow thinking mode was autonomously activated by \modelname{} (Adaptive) on the MATH500 benchmark, categorized by difficulty level. This illustrates the model's ability to adapt reasoning depth to task complexity.}
    \label{fig:math500-fsa}
\end{figure}

In addition to evaluating overall accuracy and average token usage, we further investigate the relationship between the observed efficiency gains and the model's ability to adaptively switch between fast and slow thinking modes based on problem difficulty. Specifically, we analyze the proportion of problems for which the model autonomously adopts the slow thinking mode. Our findings indicate that this proportion varies significantly with task complexity: For the relatively simpler GSM8K dataset, slow thinking mode usage drops to 14.56\%. On the MATH500 benchmark, as illustrated in Figure~\ref{fig:math500-fsa}, the tendency to engage slow thinking mode increases monotonically with the problem's difficulty level. These findings strongly suggest that our adaptive \modelname{} is capable of efficiently allocating reasoning resources by favoring fast, concise solutions for easier problems while automatically switching to more detailed, slow reasoning for harder ones, thereby effectively balancing computational efficiency and reasoning accuracy.

\begin{table}[tb]
    \centering
    \renewcommand{\arraystretch}{1.2}
    \caption{Accuracy of \modelname{} (SFT) on AIME 2024 across different SFT iterations.} 
    \label{tab:iteration}
    \begin{tabular}{@{}l c c c @{}}
    \toprule
    \textbf{Model}         & \textbf{Iter1} & \textbf{Iter2} & \textbf{Iter3} \\
    \midrule
    \modelname{} (SFT)   & 50.42          & 53.75          & 57.25          \\
    \bottomrule
    \end{tabular}
\end{table}

\subsection{Ablation Studies on SFT Strategy}

\paragraph{Data Selection based on Complexity Score.}

\begin{figure}[t!]
    \centering
\includegraphics[width=\textwidth]{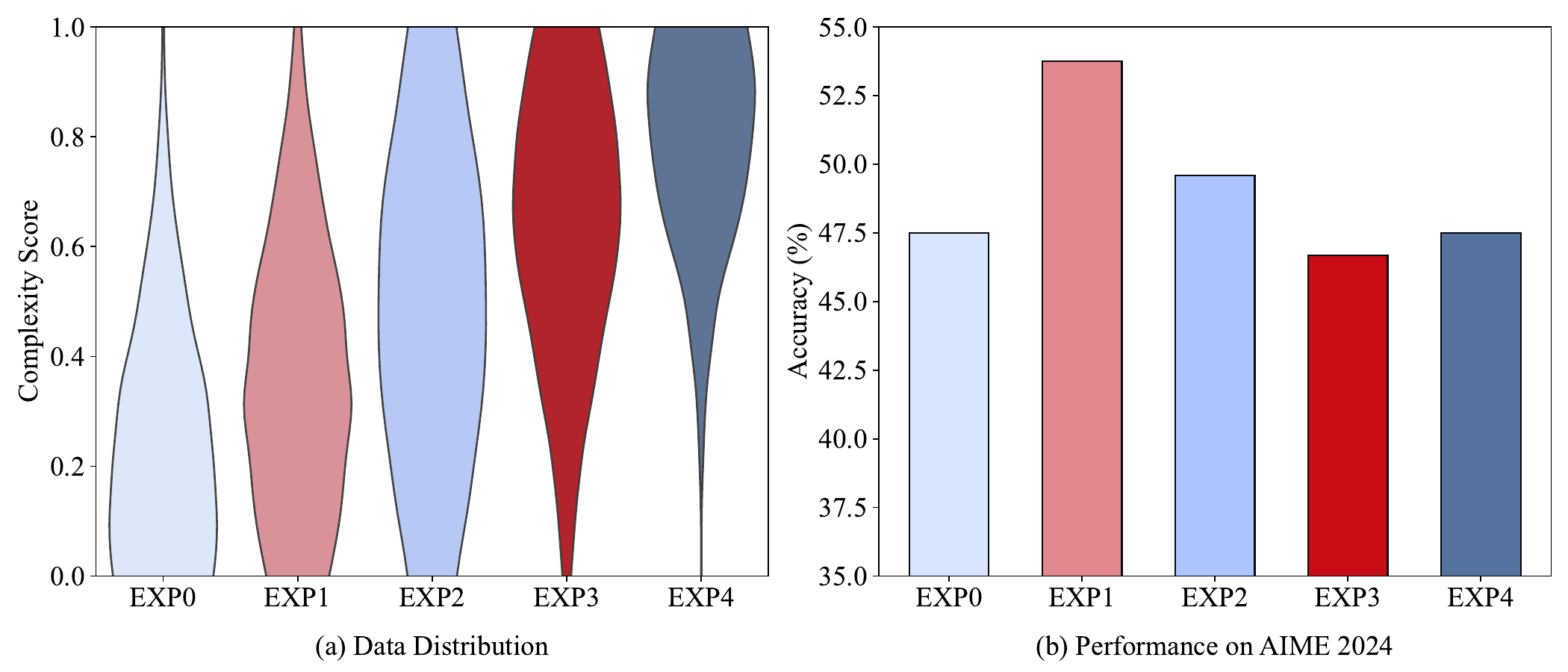}
    \caption{Experiment results on AIME 2024 when using 5 different sets of training data in terms of complexity score. }
    \label{fig:diff_complexity}
\end{figure}

In our iterative training process, we utilize Eq.~\eqref{eq:select-prob} to select data based on their complexity scores, with a focus on medium leaning towards simpler complexity data. To empirically evaluate the impact of this selection strategy, we conduct five experiments during the second iteration, labeled EXP0 to EXP4. In these experiments, we progressively increase the $\mu$ value in Eq.~\eqref{eq:select-prob}. Notably, EXP0 excludes highly complex samples, while EXP4 excludes overly simple ones. The data distribution and performance on AIME 2024 are illustrated in Figure~\ref{fig:diff_complexity}.

Figure~\ref{fig:diff_complexity} (b) shows that the model trained with questions of medium leaning toward easier complexity (i.e., EXP1) successfully outperforms the others trained high complexity  questions (i.e., EXP2, EXP3 \& EXP4) and low complexity questions (i.e., EXP0).
This suggests that the medium complexity question can effectively elicit the potential reasoning capability in the base model.

Morever, both EXP0 and EXP4, which represent the two extremes of the complexity spectrum---entirely easy and entirely hard samples respectively---exhibit noticeable performance degradation. This suggests that removing either end of the complexity spectrum leads to a suboptimal performance and limits the model ability to generalize. 
These findings highlight the importance of maintaining a balanced and comprehensive distribution of data complexities during distillation. 
In particular, completely excluding extremely hard samples (those with zero success rate) in EXP0 prevents the model from acquiring more advanced capabilities or novel behavioral patterns. This observation suggests that gradually incorporating a limited number of unmastered, high-complexity samples during SFT can contribute to expanding the model capability boundaries.
While progressively improving the model's performance, it is important to ensure that the model is consistently exposed to new knowledge and tasks.

\paragraph{Iteration Effects}
We present the results on AIME 2024 of our \modelname{} (SFT version) after each iteration of the supervised fine-tuning process in Table~\ref{tab:iteration}. As shown, the performance improves significantly with each iteration, highlighting the effectiveness of our model-aware iterative training approach. Subsequent iterations beyond the third yielded only marginal improvements, leading us to conclude the process after three iterations for this set of experiments.

\begin{table}[tb]
    \centering
    \renewcommand{\arraystretch}{1.2}
    \caption{Performance comparison of \modelname{} (SFT) with and without inter-iteration Model Merging (detailed in Section~\ref{sec:inter_iteration_model_merging}).} 
    \label{tab:ModelMerge}
    \begin{tabular}{@{}lccc@{}}
    \toprule
    \textbf{Model Configuration}            & \textbf{AIME 2024} & \textbf{LiveCodeBench} & \textbf{GPQA Diamond} \\ 
    \midrule
    \modelname{} (SFT, without Merging) & 54.58              & 40.44                  & 52.53                 \\
    \modelname{} (SFT, with Merging)    & 57.25              & 44.48                  & 53.53                 \\ 
    \bottomrule
    \end{tabular}
\end{table}

\paragraph{Model Merging Effects} 
As discussed in Section~\ref{sec:inter_iteration_model_merging}, we leverage an inter-iteration model merging technique to mitigate potential issues such as declining general capabilities or catastrophic forgetting during the iterative SFT training. Table~\ref{tab:ModelMerge} shows the performance of the \modelname{} (SFT version) model with and without this model merging strategy. Across three representative reasoning benchmarks, namely AIME 2024 for mathematics, LiveCodeBench for coding, and GPQA Diamond for general language reasoning, a consistent performance gain is observed when applying model merging. This indicates the importance of leveraging and consolidating model parameter variations that arise during the dynamic process of iterative training.

\begin{table}[tb] 
\centering
\renewcommand{\arraystretch}{1.2}
\caption{Comparison of Baseline versus Repetition Self-repair on the MATH500 benchmark, results averaged over 8 seeds. ``Acc'' is accuracy. ``Repeat'' denotes the average number of samples exhibiting significant repetitive content. ``Truncation'' indicates the average count of samples truncated due to exceeding maximum context length, often exacerbated by repetition. ``Token Len'' is the average output token length.}
\label{math500_repeat_res_tab} 
\small 
\begin{tabular}{@{}lcccc@{}}
\toprule
\textbf{Method}             & \textbf{Acc (\%)} & \textbf{Repeat} & \textbf{Truncation} & \textbf{Token Len}  \\ 
\midrule
Baseline (SFT)                   & 84.8              & 12.25           & 12.63               & 2422.8              \\
Repetition Self-repair      & 85.3              & 2.25            & 2.38                & 2340.2              \\ 
\bottomrule
\end{tabular}
\end{table}

\subsection{Ablation Studies on Repetition Self-repair}
\label{subsec:ablation_repetition_self_repair}

We conducted comparative experiments on the MATH500 benchmark, using 8 different random seeds, to systematically evaluate the impact and efficiency of our Repetition Self-repair strategy (detailed in Section~\ref{sec:repetition_self_repair}). In these experiments, the n-gram size for detection was set to 512, the local window length to 1024 tokens, and the Jaccard Similarity threshold to 0.6. Repetition detection was performed every 2048 decoded tokens (\ie, $t_{\text{detect}}=2048$).

The experimental results are presented in Table~\ref{math500_repeat_res_tab}. First, the accuracy of our method (\modelname{} with Repetition Self-repair) shows an improvement of approximately 0.5 percentage points compared to the Baseline, suggesting a slight enhancement in reasoning ability or answer correctness due to more coherent outputs. Second, and more significantly, our method drastically reduces the average number of samples containing repetitive content from 12.25 to 2.25, and the average number of samples truncated due to excessive length (often caused by repetition) from 12.63 to 2.38. Correspondingly, the average length of the model output also decreases. 
These results demonstrate that the problem of models repeatedly outputting content is effectively alleviated by our strategy. Furthermore, as illustrated in Figure~\ref{fig:self_repair_repeat_overall}, when the self-repair mechanism detects repetitive sentences (conceptually highlighted in red), it inserts guiding prompts (blue), prompting the model to reflect on the repetitive output (green) and subsequently continue reasoning in a correct and non-repetitive manner.

\begin{figure}[htb]
    \centering
    \includegraphics[width=1\linewidth]{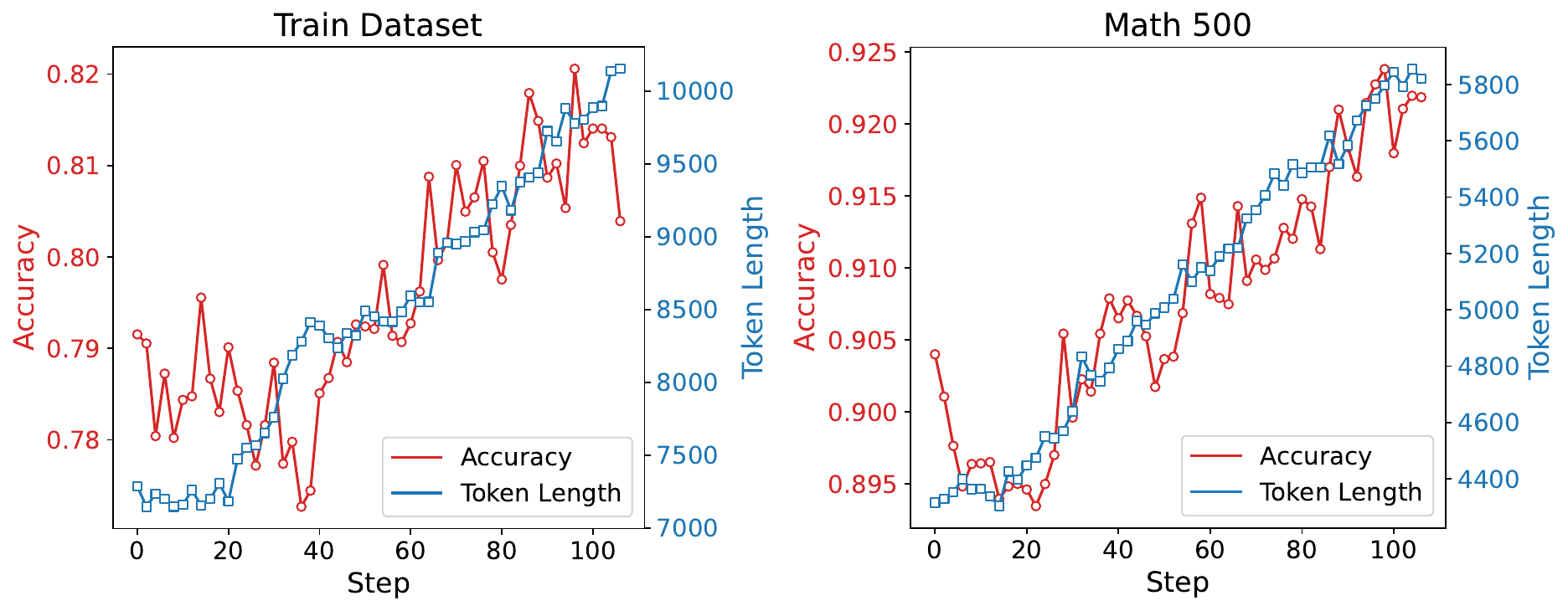}
    \caption{Illustrative RL training progression for mathematical reasoning tasks, starting from an early SFT checkpoint on Ascend NPUs. The y-axis represents a key performance metric (e.g., training accuracy or reward), while the x-axis represents training steps. This curve demonstrates stable learning and capability improvement during the initial RL validation phase.}
    \label{fig:rl_training}
\end{figure}

\begin{figure}[t]
\centering
\begin{minipage}[t]{0.49\textwidth}
    \centering
    \includegraphics[width=\linewidth]{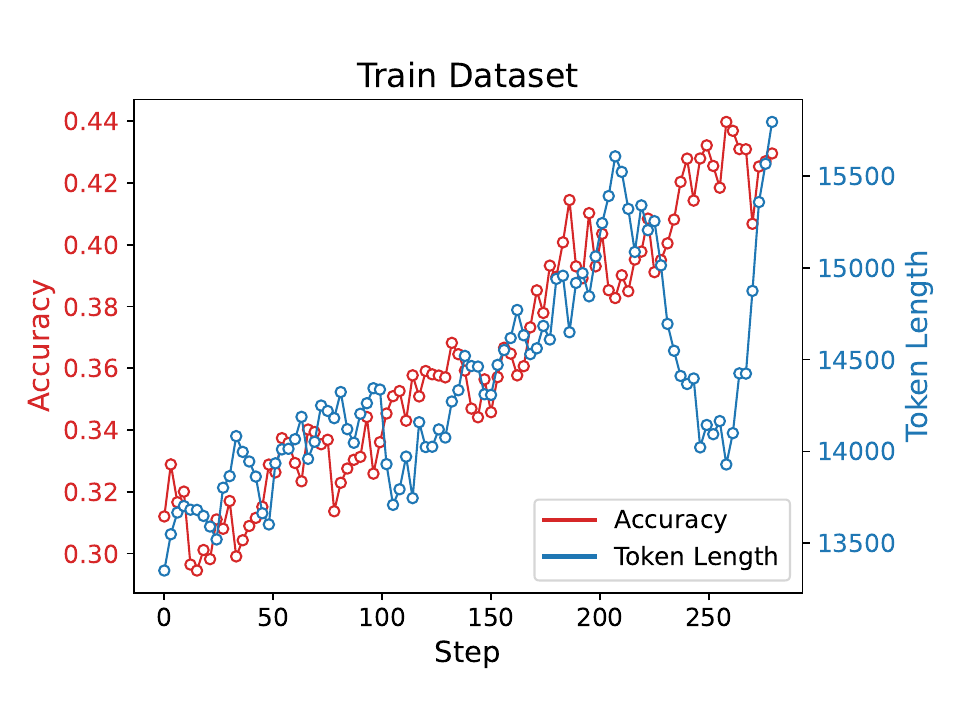}
    \label{fig:training}
\end{minipage}
\hfill
\begin{minipage}[t]{0.49\textwidth}
    \centering
    \includegraphics[width=\linewidth]{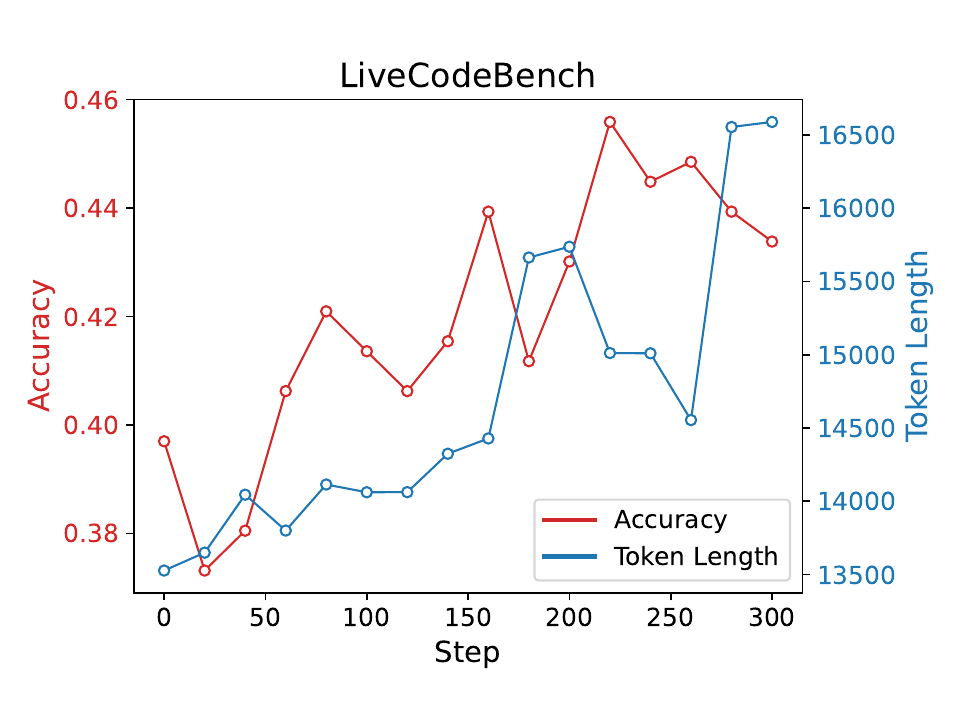}
    \label{fig:accuracy}
\end{minipage}
\vspace{-1em}
\caption{The learning and test of code generation tasks by RL on Ascend NPUs. The y-axis indicates a relevant performance metric (e.g., pass rate or reward), and the x-axis denotes training steps. This figure highlights the model's consistent improvement during the initial RL validation for code generation.}
\label{fig:rl_code_training}
\end{figure}

\subsection{Ablation Studies on  Reinforcement Learning Strategy}
\label{sec:rl_validation}

To validate and refine our RL approach, particularly its stability and efficacy on Ascend NPUs, we initially conducted a series of experiments using an early version of our SFT model. These preliminary studies, focusing on mathematical reasoning and code generation, provided crucial insights and demonstrated the viability of long, stable RL training runs before proceeding to train our final \modelname{} with the complete RL methodology.

\paragraph{Initial RL Validation on Mathematical Reasoning.}
For the mathematical reasoning domain, we first investigated the training stability of our RL setup. Starting with an early slow-thinking SFT model checkpoint, which is similar with ``cold start'', we curated a model-aware training set of approximately 30,000 samples from the ORZ Math dataset, employing the data filtering mechanism detailed in Section~\ref{sec:curriculum_data_mixing}. The RL training process on Ascend chips, as depicted in Figure~\ref{fig:rl_training}, exhibited notable stability. Following an initial warm-up phase, the training accuracy demonstrated a steady increase. More importantly, evaluations on challenging benchmarks such as MATH500 revealed consistent improvements in both average response length and task accuracy, confirming a stable and positive learning trajectory for our RL approach in this domain.

\paragraph{Initial RL Validation on Code Generation.}
Similarly, for code generation, we conducted RL training using an early SFT model. We selected 16,384 prompts from competitive programming platforms like Codeforces and AtCoder, specifically choosing problems with a baseline pass@1 rate greater than 0 and less than 1 to focus on areas where the model had potential for improvement. These training samples encompassed both function-calling and stdin/stdout formats and covered a diverse range of algorithmic domains. During this RL training phase, we employed a clip-higher mechanism with $\epsilon_{high}=0.28$ to maintain entropy stability. The training and testing curves, illustrated in Figure~\ref{fig:rl_code_training}, show a clear positive trend. Notably, this preliminary RL training resulted in a $\mathbf{6\%}$ absolute improvement on the LiveCodeBench benchmark with only 300 steps.

\paragraph{Transition to Final Model Training with MARS.}
The successful and stable learning demonstrated in these initial RL experiments on both mathematical reasoning and code generation using an early SFT model on Ascend NPUs provided strong evidence for the robustness of our RL training pipeline. Encouraged by these findings, we proceeded to apply our comprehensive RL strategy, featuring the Multi-Source Adaptive Reward System (MARS) as detailed in Section~\ref{sec:mars_rl}, to our final, optimized SFT model. This subsequent phase of long and stable RL training, guided by MARS, resulted in the development of the final \modelname{} whose performance is reported in this work.

\subsection{Extension to Domain-specific Tasks}
\label{subsec:domain_specific_extension}

The \modelname{} model, along with its comprehensive training protocol, establishes a solid foundation for general-domain reasoning capabilities. To further explore the adaptability of our system and its potential for acquiring specialized industry-level reasoning power, we undertook a systematic effort in domain-specific adaptation. We selected the legal domain as a proof-of-concept for this extension, given its significant reliance on sophisticated, human-like reasoning for accurate analysis of complex legal matters.

\subsubsection{Domain-specific Data Preparation} 
\label{ssubsec:domain_data}

\paragraph{Data Source and Composition}
To extend the reasoning power of \modelname{} to a specialized domain like law, it is crucial to distill domain-specific chain-of-thoughts that accurately reflect expert reasoning processes. To generate these data, we first collected multiple source datasets from a variety of open-source and in-house SFT data collections, spanning numerous industries but with a strong emphasis on legal domains. Both Chinese and English data sources were considered. The dataset collection strategy prioritized both the nature of the industry and the diversity of tasks represented. Initial data included general industry scenarios to provide a foundational logic base for business contexts, covering single-turn, multi-turn, and reasoning-based question-answering settings. To specifically enhance performance in our target domain, we then introduced a significant volume of legal data, focusing on the analysis of laws, case files, and contracts—tasks central to a lawyer's routine work.

\paragraph{Data Processing and Quality Control}
The curated domain-specific data underwent several processing steps. First, regarding quality control, we noted that meticulous removal of duplicate queries across all source datasets within the domain data is essential to avoid redundancy and enhance training efficacy. A rigorous de-duplication process was implemented using a combination of exact and fuzzy matching techniques to prevent the model from being unduly biased towards certain entries or question types.

Second, concerning chain-of-thought data distillation and filtering, we adopted a knowledge distillation approach for the majority of the datasets. While the queries in these datasets were generally considered valuable for their reflection of business or legal requirements, the quality of the original chain-of-thoughts and answers varied. Thus, these were often replaced by new solutions distilled from more advanced large language models. Furthermore, any chain-of-thought data identified by our in-house automatic inspection pipeline as indicative of over-thinking or containing repetitive output was discarded or flagged for revision. The final reasoning data for domain adaptation was calibrated to an approximate entry ratio of 1:2 for general industry versus legal domain content. This domain-specific data was then further mixed with the general post-training data of \modelname{} to constitute the final training set for domain adaptation.

\begin{figure}[htb] 
    \centering
    \includegraphics[width=1\linewidth]{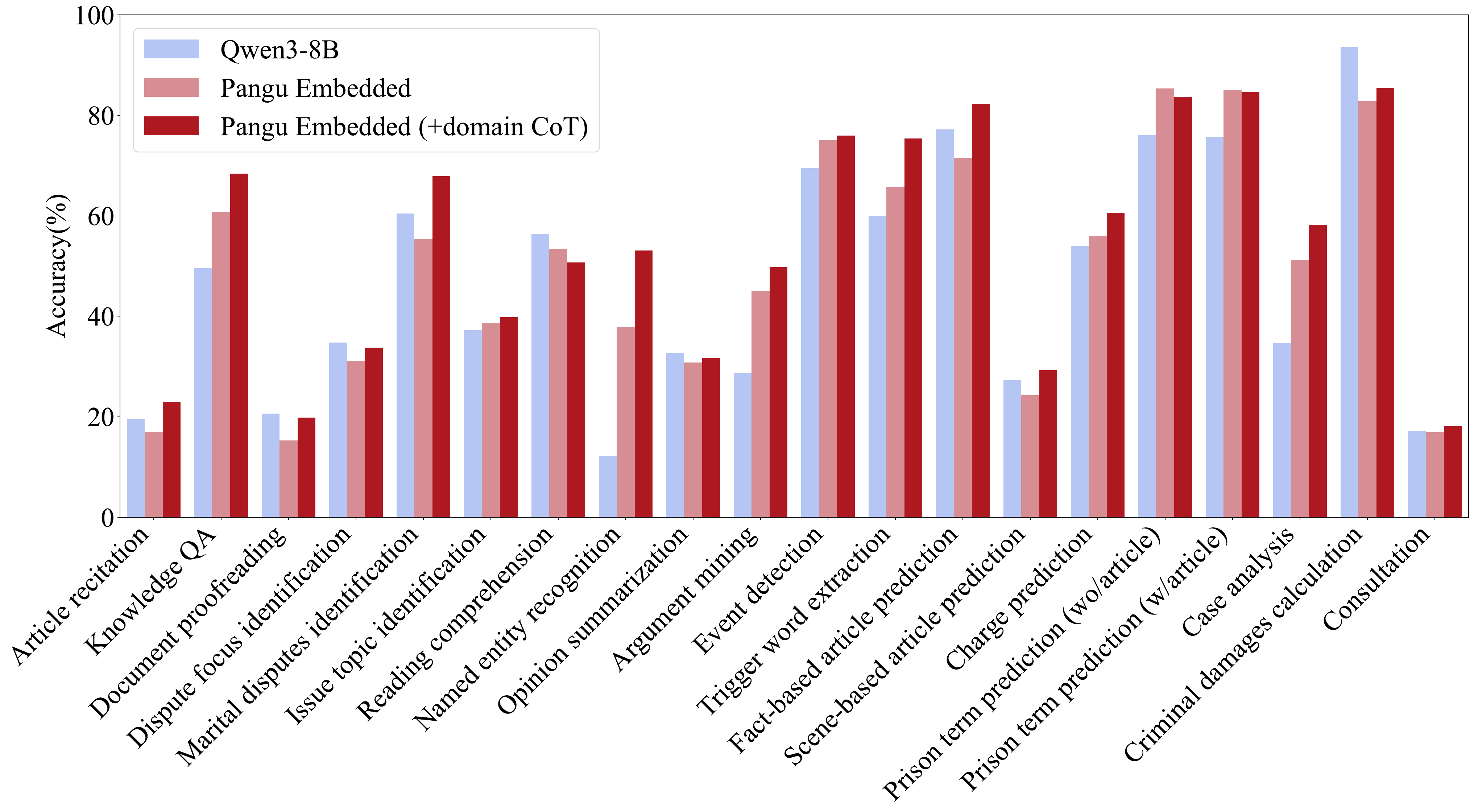} 
    \caption{Performance comparison of different reasoning models on selected tasks from the LawBench benchmark~\cite{fei2023lawbench}.}
    \label{fig:legal_performance}
\end{figure}

\subsubsection{Results in the Legal Domain} 
\label{ssubsec:domain_results}

\paragraph{Benchmark}
To assess the model's performance in the legal domain, we employed LawBench~\cite{fei2023lawbench}, a comprehensive benchmark for legal intelligence. LawBench consists of 20 distinct sub-tasks which can be categorized, based on increasing task difficulty, into legal knowledge memorization, legal understanding, and legal application.

\paragraph{Evaluation on Domain Tasks}
Our systematic evaluation on LawBench, as partially illustrated in Figure~\ref{fig:legal_performance}, reveals the impact of domain-specific adaptation. Before domain-specific training, on the 20 legal tasks of LawBench, the average accuracy of a strong baseline, Qwen3-8B (Thinking mode), was 46.88\%. In comparison, our general-domain \modelname{} (slow thinking mode) achieved an average accuracy of 49.97\%, surpassing the baseline by 3.09 percentage points.
After undergoing domain-specific training with the curated legal data, \modelname{} exhibited a significant gain in legal knowledge and reasoning capabilities. Its performance improved on 17 out of the 20 tasks, leading to a new average accuracy of 54.59\% on LawBench. This represents a further improvement of 4.62 percentage points over its general-domain version. These promising results indicate substantial room for enhancing performance on specialized tasks through targeted post-training, building upon a strong general reasoner. The observed advantage of \modelname{} over Qwen3-8B, both before and after domain adaptation, might also be partially attributed to differences in their respective general pre-training or initial fine-tuning data mixtures.

In particular, we manually inspected the legal chain-of-thought generated for challenging case analysis tasks, which require robust legal knowledge application and reasoning. As qualitatively illustrated by an example in Figure~\ref{fig:case}, while multiple models might arrive at the correct final answer, the domain-adapted \modelname{} demonstrated significantly more stringent and accurate reasoning. This was manifested, for instance, by its correct application of principles related to the burden of proof under the Civil Code of the People's Republic of China, where other models might falter. Such examples support the conclusion that domain adaptation improves not only accuracy but also the logical quality and precision of the reasoning process.

\section{Related Work}

\textbf{Reasoning Models.} The advent of OpenAI's o1~\cite{openai2024o1} marked a significant milestone, demonstrating the substantial potential of reasoning models to excel in diverse tasks by leveraging extended chains of thought. Subsequent advancements, exemplified by models such as DeepSeek-R1~\cite{guo2025deepseek} and QwQ~\cite{qwq2024}, have further elucidated the underlying algorithms for constructing powerful reasoning models, thereby providing the research community with clearer algorithmic roadmaps. However, a considerable portion of these works has not fully emphasized the criticality of the training data recipe, an indispensable component for ensuring robust and generalizable model performance. While some recent studies~\cite{muennighoff2025s1,ye2025limo} have reported notable performance gains using remarkably small supervised fine-tuning (SFT) datasets (e.g., around 1,000 examples) for reasoning tasks, such methodologies may not yield optimal results for other foundation models due to inherent variations in their architectural designs and capabilities.

\textbf{RL Infrastructure for LLMs.} Recent progress in reinforcement learning infrastructure tailored for LLMs has predominantly centered on developing scalable training frameworks and effectively integrating human feedback. Ouyang et al.~\cite{ouyang2022training} introduced pioneering techniques for fine-tuning LLMs via Reinforcement Learning from Human Feedback (RLHF), capitalizing on distributed systems to achieve efficient training. Concurrently, Bai et al.~\cite{bai2022training} proposed Constitutional AI, a novel approach that synergizes RL with rule-based constraints to guide model behavior in accordance with predefined ethical guidelines. Foundational algorithmic contributions, such as Proximal Policy Optimization (PPO) established by Schulman et al.~\cite{schulman2017proximal}, have been widely adopted within RL-based LLM training pipelines. Furthermore, distributed RL frameworks like Ray RLlib~\cite{liang2018rllib} have been adapted to manage the substantial computational demands associated with LLM fine-tuning. Nevertheless, the majority of existing practices are primarily architected for GPU-based clusters. Consequently, there is a pressing need for optimization efforts to enhance RL infrastructures, particularly for Ascend-based cluster environments.

\textbf{Knowledge Distillation.} Knowledge distillation~\cite{xu2024survey,yang2024survey} has emerged as a prominent technique for transferring knowledge or capabilities from a typically larger or more proficient teacher model to a student model. Early practices in this domain explored distillation by aligning the output distributions of both models~\cite{sreenivas2024llm}. However, the applicability of such methods is contingent upon both models sharing an identical vocabulary set. More recent investigations~\cite{AM-DeepSeek-R1-Distilled-1.4M,huang2024o1,yu2024distilling} have successfully employed knowledge distillation to imbue models with strong reasoning abilities by distilling extensive CoT samples from advanced teacher models like DeepSeek-R1. To further enhance the quality of these distilled samples, techniques such as Monte Carlo Tree Search (MCTS)~\cite{yin2025towards} have been incorporated. In this paper, our focus is on straightforward yet effective distillation methodologies. We further aim to explore the nuanced impact of data strategies on the development of high-performing LLMs, taking into account the inherent capabilities of the student model.

\textbf{Model-aware Training.} Initial efforts in the post-training of LLMs predominantly concentrated on curating high-quality training data from a general standpoint, often without explicit consideration of model-specific characteristics or potential issues~\cite{liumakes,zhou_lima_2023}. While this universal approach has yielded improvements in model performance, recent studies have indicated that data distributions deviating significantly from a base model's learned representations can be challenging for the model to assimilate and may, in some instances, lead to performance degradation~\cite{ren_learning_2024}. Consequently, an increasing body of research advocates for model-specific data selection strategies~\cite{du2023mods,li-etal-2024-quantity}. These studies, however, have not extensively explored distillation scenarios. Our work extends this concept by proposing and integrating an iterative distillation pipeline that leverages a novel model-aware complexity score.

\textbf{Fast and Slow Thinking.} The integration of ``fast'' (intuitive, heuristic) and ``slow'' (deliberative, reasoning-intensive) cognitive paradigms into large reasoning models has recently garnered significant attention as a means to enhance both the quality of reasoning and operational efficiency. There are several concurrent works in this field. The FAST framework~\cite{xiao2025fast}, for instance, dynamically adapts the depth of reasoning based on the input question's characteristics, employing model-based metrics for question assessment, an adaptive thinking reward mechanism, and difficulty-aware KL regularization. Similarly, AutoThink~\cite{tu2025learning} introduces a multi-stage RL framework that enables R1-style LLMs to acquire adaptive reasoning behaviors. This is achieved by utilizing an ellipsis prompt to stochastically switch between thinking and non-thinking modes, thereby allowing the model to allocate reasoning effort dynamically. AdaptThink~\cite{zhang2025adaptthink}, another novel RL algorithm, trains reasoning models to select the optimal thinking mode adaptively, guided by problem difficulty, through a constrained optimization objective and an importance sampling strategy. These approaches collectively contribute to the evolving paradigm of adaptive reasoning in large models, each offering unique methodological insights for balancing the trade-off between rapid responses and thorough deliberation. Furthermore, recent technical reports~\cite{yang2025qwen3, bercovich2025llama} have also proposed mechanisms for manually switching between fast and slow thinking modes. Both Llama-Nemotron~\cite{bercovich2025llama} and Qwen3~\cite{yang2025qwen3} aim to strike an effective balance between reasoning depth and computational efficiency, thereby advancing the capabilities of reasoning models by providing flexible and resource-aware solutions for a diverse array of tasks.

\section{Conclusion and Discussion}
\label{sec:conclusion}
This work introduced \modelname{}, an efficient LLM reasoner developed on Ascend NPUs, which achieves state-of-the-art reasoning accuracy among similarly-sized models. Our core contribution is a two-stage training framework. Stage 1 constructs a robust basic reasoner through iterative distillation incorporating model-aware data complexity selection, inter-iteration checkpoint merging for knowledge consolidation, and large-scale reinforcement learning optimized with a latency-tolerant scheduler and the multi-source adaptive reward system. Stage 2 further endows \modelname{} with a novel dual-system fast-slow thinking capability, featuring both user-controlled manual switching and adaptive mode selection to dynamically balance reasoning depth with computational efficiency. Complemented by a repetition self-repair mechanism for improved generation quality, our work demonstrates a practical path towards developing compact, yet powerful and deployable LLM reasoners.

Our study also offers insights into effective data strategies for training reasoning LLMs, notably the value of tailoring data complexity to the training paradigm—medium, simpler data for SFT-based distillation, and broader, higher-complexity data for RL-based refinement. This points towards promising future directions, such as developing hybrid SFT-RL frameworks that leverage model-aware complexity scheduling for more dynamic and efficient knowledge transfer and policy refinement. We believe this work provides a foundational approach for creating more computationally sustainable and powerful reasoning models.

\bibliographystyle{plain}
\bibliography{ref}  

\clearpage
\appendix
\section{Contributions and Acknowledgments}
\noindent\textbf{Core Contributors}
Hanting Chen, Yasheng Wang, Kai Han, Dong Li, Lin Li, Zhenni Bi, Jinpeng Li, Haoyu Wang, Fei Mi, Mingjian Zhu, Bin Wang, Kaikai Song, Yifei Fu, Xu He, Yu Luo, Chong Zhu, Quan He, Xueyu Wu, Wei He, Hailin Hu, Yehui Tang, Dacheng Tao, Xinghao Chen, Yunhe Wang

\textbf{Contributors}
Binwei Yan, Can Chen, Chan Tsz Ho, Chen Zhong, Chenyi Pan, Chuanjian Liu, Fanyi Du, Fisher Yu, Heyuan Deng, Huiling Zhen, Liangjun Feng, Lin Yang, Luocheng Hu, Nianzu Zheng, Qianyi Sun, Qiang Gu, Ruiming Tang, Shixiong Kai, Shuo Han, Tianyu Guo, Ting Hu, Tong Teng, Weiwen Liu, Wulong Liu, Xianzhi Yu, Xiaojun Meng, Xing Li, Xingshan Zeng, Xinyi Dai, Yan Xu, Yihan Hu, Yinfei Pan, Ying Nie, Yingxue Zhang, Yufei Wang, Yubin Wang, Yuqi Cui, Yunsheng Ni, Zelin Chen, Zhao Liu, Zheyuan Bai, Zhicheng Liu, Zhenhang Weng, Zixuan Yue, Zongyuan Zhan

\clearpage
\section{Response Examples}

\begin{figure}[H]
    \centering
    \includegraphics[width=0.92\linewidth]{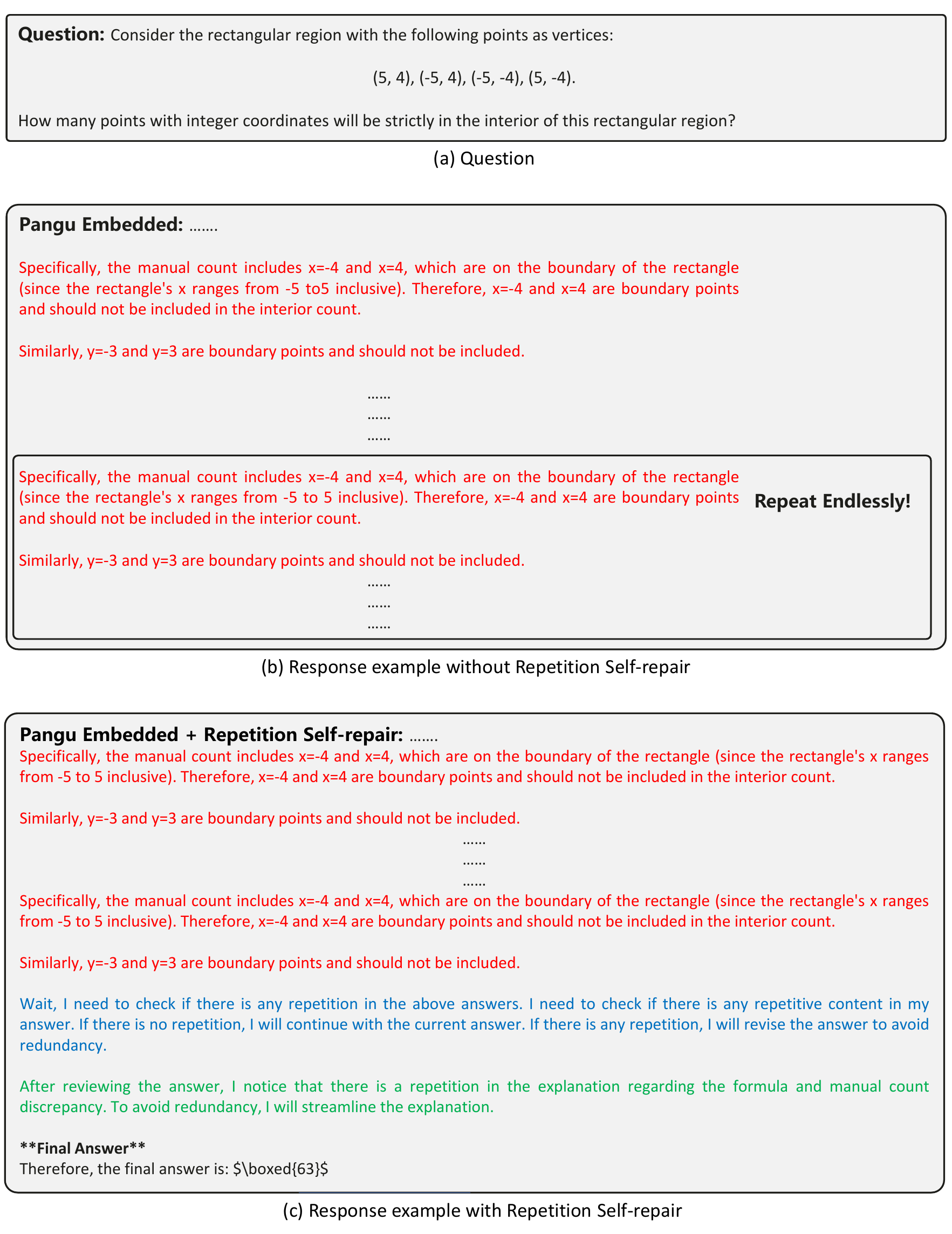} 
    \caption{A demonstration of output from \modelname{} with Repetition Self-repair on a MATH benchmark task. (a) Shows the initial query. (b) Illustrates how a model without self-repair can get stuck in repetitive paragraphs. (c) Shows the Repetition Self-repair mechanism in action: detected repetitive sentences are notionally marked (e.g., red highlight in internal representation), the guiding prompt (e.g., blue) is inserted, leading the model to reflect on and correct the repetition (e.g., green), ultimately producing a correct and non-repetitive reasoning path.}
    \label{fig:self_repair_repeat_overall}
\end{figure}
\begin{figure}[htb]
    \centering
    \includegraphics[width=1\linewidth]{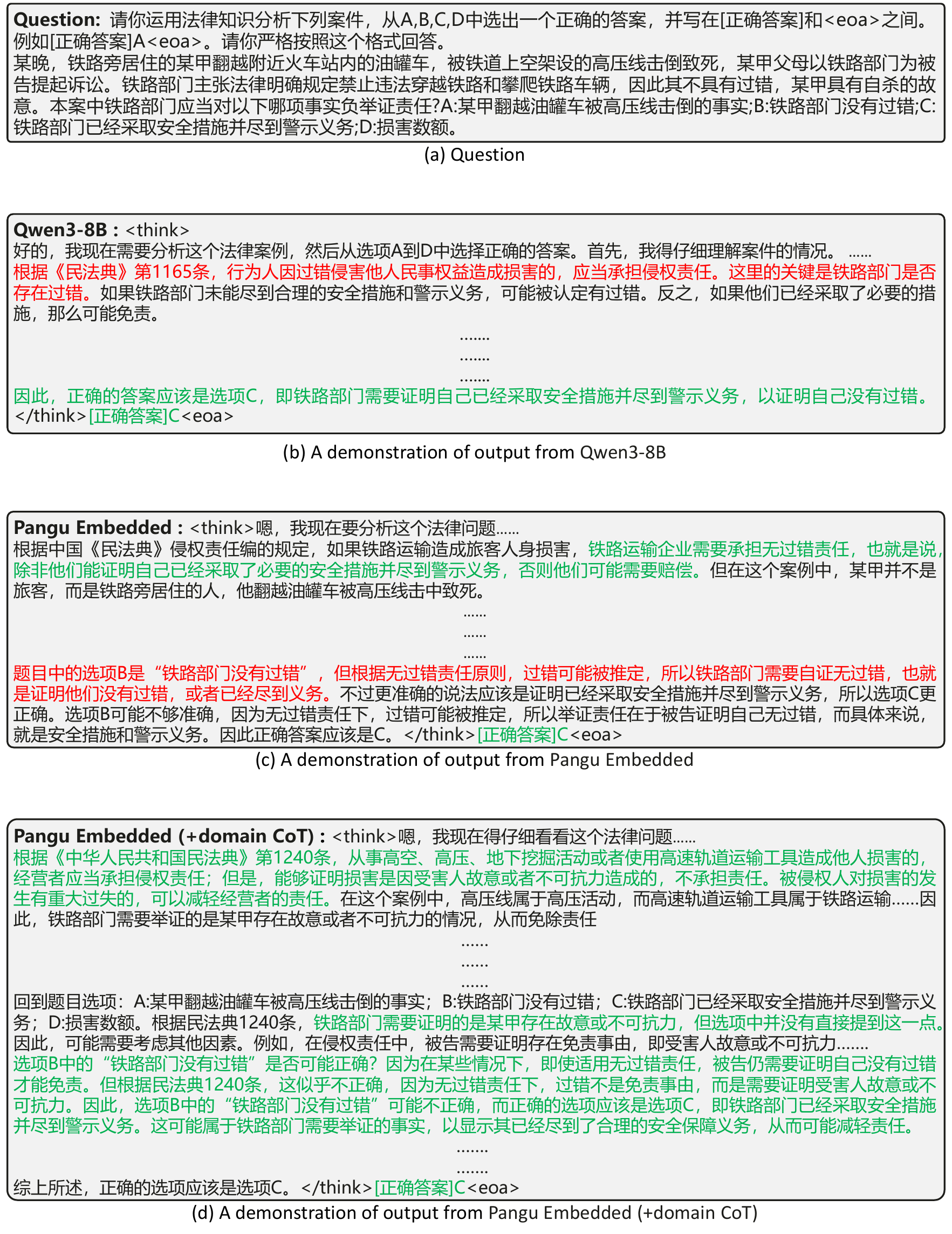}
    \caption{A demonstration of output from different reasoning models under China's law. Incorrect reasoning is labeled as red, and correct reasoning is highlighted as green.}
    \label{fig:case}
\end{figure}


\end{document}